\documentclass{article} 
\usepackage{iclr2023_conference_latex,times}

\usepackage{amsmath,amsfonts,bm}

\def\eqref#1{equation~\ref{#1}}

\def\1{\bm{1}}

\DeclareMathAlphabet{\mathsfit}{\encodingdefault}{\sfdefault}{m}{sl}
\SetMathAlphabet{\mathsfit}{bold}{\encodingdefault}{\sfdefault}{bx}{n}

\usepackage{hyperref}
\usepackage{url}
\usepackage{xspace}
\usepackage{graphicx}
\usepackage{caption}
\usepackage{multirow}
\usepackage{amsmath}
\usepackage{amssymb}
\usepackage{booktabs}
\usepackage{times}
\usepackage{epsfig}
\usepackage{enumitem}
\usepackage[misc]{ifsym}
\usepackage{makecell}
\usepackage{cases}
\usepackage{xcolor}
\usepackage{wrapfig}

\newcommand{\nickname}{EVA3D}

\title{\nickname{}: Compositional 3D Human Generation from 2D Image Collections}

\author{Fangzhou Hong,\quad Zhaoxi Chen,\quad Yushi Lan,\quad Liang Pan,\quad Ziwei Liu\textsuperscript{~\Letter} \\
S-Lab, Nanyang Technological University\\
\texttt{\small{\{fangzhou001, zhaoxi001, yushi001, liang.pan, ziwei.liu\}@ntu.edu.sg}} \\
}

\newcommand{\ie}{\emph{i.e.}\xspace}
\newcommand{\eg}{\emph{e.g.}\xspace}

\iclrfinalcopy 
\begin{document}

\onecolumn{%
\renewcommand\twocolumn[1][]{#1}%
\maketitle

\newcommand\extrafootertext[1]{%
    \bgroup
    \renewcommand\thefootnote{\fnsymbol{footnote}}%
    \renewcommand\thempfootnote{\fnsymbol{mpfootnote}}%
    \footnotetext[0]{#1}%
    \egroup
}
\extrafootertext{\textsuperscript{\Letter}~Corresponding author}

\begin{center}
    \centering
    \captionsetup{type=figure}
    \vspace{-15pt}
    \includegraphics[width=\textwidth]{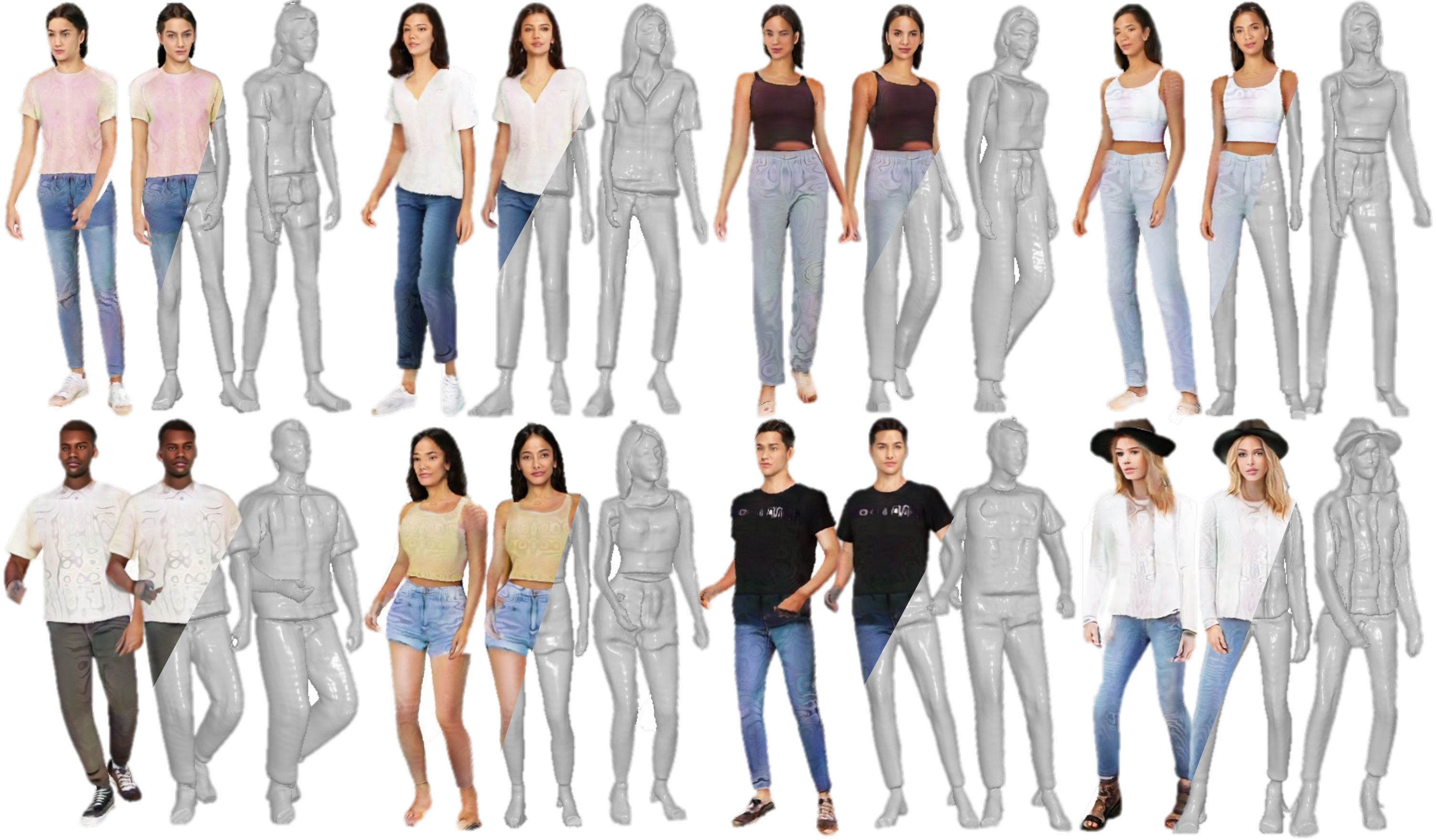}
    \vspace{-20pt}
    \captionof{figure}{\nickname{} generates high-quality and diverse 3D humans with photo-realistic RGB renderings and detailed geometry. Only 2D image collections are used for training.
    }
    \label{fig:teaser}
\end{center}%
}

\begin{abstract}

Inverse graphics aims to recover 3D models from 2D observations.
Utilizing differentiable rendering, recent 3D-aware generative models have shown impressive results of rigid object generation using 2D images.
However, it remains challenging to generate articulated objects, like human bodies, due to their complexity and diversity in poses and appearances.
In this work, we propose, \textbf{\nickname{}}, an unconditional 3D human generative model learned from 2D image collections only. 
\nickname{} can sample 3D humans with detailed geometry and render high-quality images (up to $512\times 256$) without bells and whistles (\eg super resolution). 
At the core of \nickname{} is a compositional human NeRF representation, which divides the human body into local parts. Each part is represented by an individual volume. 
This compositional representation enables \textbf{1)} inherent human priors, \textbf{2)} adaptive allocation of network parameters, \textbf{3)} efficient training and rendering.
Moreover, to accommodate for the characteristics of sparse 2D human image collections (\eg imbalanced pose distribution), we propose a pose-guided sampling strategy for better GAN learning.
Extensive experiments validate that \nickname{} achieves state-of-the-art 3D human generation performance regarding both geometry and texture quality. 
Notably, \nickname{} demonstrates great potential and scalability to ``inverse-graphics'' diverse human bodies with a clean framework. Project page: \url{https://hongfz16.github.io/projects/EVA3D.html}.

\end{abstract}
\section{Introduction}

Inverse graphics studies inverse-engineering of projection physics, which aims to recover the 3D world from 2D observations. 
It is not only a long-standing scientific quest, but also enables numerous applications in VR/AR and VFX. 
Recently, 3D-aware generative models~\citep{pigan, stylesdf, eg3d, GRAM} demonstrate great potential in inverse graphics by learning to generate 3D rigid objects (\eg human/animal faces, CAD models) from 2D image collections. 
However, human bodies, as articulated objects, have complex articulations and diverse appearances. Therefore, it is challenging to learn 3D human generative models that can synthesis animatable 3D humans with high-fidelity textures and vivid geometric details.
To generate high-quality 3D humans, we argue that two main factors should be properly addressed: \textbf{1) 3D human representation}; \textbf{2) generative network training strategies}.
Due to the articulated nature of human bodies, a desirable human representation should be able to explicitly control the pose/shape of 3D humans.
An articulated 3D human representation need to be designed, rather than the static volume modeling utilized in existing 3D-aware GANs.
With an articulated representation, a 3D human is modeled in its canonical pose (canonical space), and can be rendered in different poses and shapes (observation space).
Moreover, the efficiency of the representation matters in high-quality 3D human generation. Previous methods~\citep{enarfgan,eg3dhuman} fail to achieve high resolution generation due to their inefficient human representations.

In addition, training strategies could also highly influence 3D human generative models. The issue mainly comes from the data characteristics.
Compared with datasets used by \citet{enarfgan} (\eg AIST~\citep{aist}), fashion datasets (\eg DeepFashion~\citep{deepfashion}) are more aligned with real-world human image distributions, making a favorable dataset choice.
AIST only has 40 dancers, which are mostly dressed in black.
In contrast, DeepFashion contains much more different persons wearing various clothes, which benefits the diversity and quality of 3D human generation.
However, fashion datasets mostly have 
\textbf{very limited human poses} (most are similar standing poses),
and \textbf{highly imbalanced viewing angles} (most are front views).
This imbalanced 2D data distribution could hinder unsupervised learning of 3D GANs, leading to difficulties in novel view/ pose synthesis.
Therefore, a proper training strategy is in need to alleviate the issue.

In this work, we propose \textbf{\nickname{}}, an unconditional high-quality 3D human generative model from sparse 2D human image collections only.
To facilitate that, we propose a compositional human NeRF representation to improve the model efficiency.
We divide the human body into 16 parts and assign each part an individual network, which models the corresponding local volume.
Our human representation mainly provides three advantages.
\textbf{1)} It inherently describes the human body prior, which supports explicit control over human body shapes and poses. 
\textbf{2)} It supports adaptively allocating computation resources. More complex body parts (\eg heads) can be allocated with more parameters.
\textbf{3)} The compositional representation enables efficient rendering and achieves high-resolution generation.
Rather than using one big volume~\citep{eg3dhuman}, our compositional representation tightly models each body part and prevents wasting parameters on empty volumes.
Moreover, thanks to the part-based modeling, we can efficiently sample rays inside local volumes and avoid sampling empty spaces.
With the compact representation together with the efficient rendering algorithm, we achieve high-resolution ($512\times256$) rendering and GAN training without using super-resolution modules, while existing methods can only train at a native resolution of $128^2$.

Moreover, we carefully design training strategies to address the human pose and viewing angle imbalance issue.
We analyze the head-facing angle distribution and propose a pose-guided sampling strategy to help effective 3D human geometry learning.
Besides, we utilize the SMPL model to leverage its human prior during training.
Specifically, we use SMPL skinning weights to guide the transformation between canonical and observation spaces, which shows good robustness to the pose distribution of the dataset and brings better generalizability to novel pose generation. 
We further use SMPL as the geometry template and predict offsets to help better geometry learning.

Quantitative and qualitative experiments are performed on two fashion datasets~\citep{deepfashion, styleganhuman} to demonstrate the advantages of \nickname{}. We also experiment on UBCFashion~\citep{ubcfashion} and AIST~\citep{aist} for comparison with prior work. Extensive experiments on our method designs are provided for further analysis.
In conclusion, our contributions are as follows:
\textbf{1)} We are the first to achieve high-resolution high-quality 3D human generation from 2D image collections;
\textbf{2)} We propose a compositional human NeRF representation tailored for efficient GAN training;
\textbf{3)} Practical training strategies are introduced to address the imbalance issue of real 2D human image collections.
\textbf{4)} We demonstrate applications of \nickname{}, \ie interpolation and GAN inversion, which pave way for further exploration in 3D human GAN.

\section{Related Work}

\noindent\textbf{3D-Aware GAN.}
Generative Adversarial Network (GAN)~\citep{goodfellow2020generative} has been a great success in 2D image generation~\citep{stylegan1, stylegan2}. Many efforts have also been put on 3D-aware generation.
\citet{hologan, platogan} use voxels, and \citet{2dganknows3d} use meshes to assist the 3D-aware generation. With recent advances in NeRF~\citep{nerf, advancesinnerf}, many have build 3D-aware GANs based on NeRF~\citep{graf, giraffe, pigan, GRAM}. To increase the generation resolution, \citet{stylenerf, stylesdf, eg3d} use 2D decoders for super resolution. Moreover, it is desirable to lift the raw resolution, by improving the rendering efficiency, for more detailed geometry and better 3D consistency~\citep{epigraf,GRAMHD}.
We also propose an efficient 3D human representation to allow high resolution training.

\noindent\textbf{Human Generation.}
Though great success has been achieved in generating human faces, it is still challenging to generate human images for the complexity in human poses and appearances~\citep{humangan, tryongan, sarkar2021style, jiang2022text2human}. Recently, \citet{styleganhuman, fruhstuck2022insetgan} scale-up the dataset and achieve impressive 2D human generation results.
For 3D human generation, \citet{chen2022gdna} generate human geometry using 3D human dataset.
Some also attempt to train 3D human GANs using only 2D human image collections.
\citet{stylepeople, 3dawaresemanticgan} use CNN-based neural renderers, which cannot guarantee 3D consistency. \citet{enarfgan} use human NeRF~\citep{noguchi2021neural} for this task, which only trains at low resolution.
\citet{eg3dhuman, avatargen} propose to increase the resolution by super-resolution, which still fails to produce high-quality results.
\citet{avatarclip} generate 3D avatars from text inputs.

\noindent\textbf{3D Human Representations.}
3D human representations serve as fundamental tools for human related tasks. \citet{smpl,smplx,garment4d} create parametric models for explicit modeling of 3D humans. To model human appearances, \citet{habermann2021real, shysheya2019textured, yoon2021pose, neuralactor} further introduce UV maps. Parametric modeling gives robust control over the human model, but less realism. \citet{npms} use implicit functions to generate realistic 3D human body shapes. Embracing the development of NeRF, the number of works about human NeRF has also exploded~\citep{neuralbody, zhao2021humannerf, animatablenerf, xu2021h, noguchi2021neural, weng2022humannerf, chen2021animatable, su2021nerf, jiang2022selfrecon, jiang2022neuman, wang2022arah}.
\citet{hcmoco} propose to learn modal-invariant human representations for versatile down-stream tasks. \citet{humman} contribute a large-scale multi-modal 4D human dataset.
Some propose to model human body in a compositional way~\citep{mihajlovic2022coap, palafox2022spams, su2022danbo}, where several submodules are used to model different body parts, and are more efficient than single-network ones.

\section{Methodology} \label{method}


\subsection{Prerequisites} \label{method:prereq}

\noindent\textbf{NeRF}~\citep{nerf} is an implicit 3D representation, which is capable of photorealistic novel view synthesis. NeRF is defined as $\{\bm{c},\sigma\} = F_{\Phi}(\bm{x}, \bm{d})$, where $\bm{x}$ is the query point, $\bm{d}$ is the viewing direction, $\bm{c}$ is the emitted radiance (RGB value), $\sigma$ is the volume density. To get the RGB value $C(\bm{r})$ of some ray $\bm{r}(t)=\bm{o}+t\bm{d}$, namely volume rendering, we have the following formulation,
$C(\bm{r}) = \int_{t_n}^{t_f}T(t)\sigma(\bm{r}(t))\bm{c}(\bm{r}(t), \bm{d})dt$ , where $T(t)=\text{exp}(-\int_{t_n}^{t}\sigma(\bm{r}(s))ds)$
is the accumulated transmittance along the ray $\bm{r}$ from $t_n$ to $t$. $t_n$ and $t_f$ denotes the near and far bounds. To get the estimation of $C(\bm{r})$, it is discretized as
\begin{equation} \label{eq:nerf_dis}
    \hat{C}(\bm{r}) = \sum_{i=1}^{N}T_i(1-\text{exp}(-\sigma_i\delta_i))\bm{c}_i \text{, where } T_i=\text{exp}(-\sum_{j=1}^{i-1}\sigma_j\delta_j)\text{, } \delta_i=t_{i+1}-t_{i}.
\end{equation}
For better geometry, \citet{stylesdf} propose to replace the volume density $\sigma(\bm{x})$ with SDF values $d(\bm{x})$ to explicitly define the surface. SDF can be converted to the volume density as
    $\sigma(\bm{x}) = \alpha^{-1}\text{sigmoid}\left(-d(\bm{x})/\alpha\right)$,
where $\alpha$ is a learnable parameter. In later experiments, we mainly use SDF as the implicit geometry representation, which is denoted as $\sigma$ for convenience.

\noindent\textbf{SMPL}~\citep{smpl}, defined as $M(\bm{\beta}, \bm{\theta})$, is a parametric human model, where $\bm{\beta}, \bm{\theta}$ controls body shapes and poses. In this work, we use the Linear Blend Skinning (LBS) algorithm of SMPL for the transformation from the canonical space to observation spaces. Formally, point $\bm{x}$ in the canonical space is transformed to an observation space defined by pose $\bm{\theta}$ as
$\bm{x}' = \sum_{k=1}^{K}w_{k}\bm{G_k}(\bm{\theta}, \bm{J})\bm{x}$,
where $K$ is the joint number, $w_{k}$ is the blend weight of $\bm{x}$ against joint $k$, $\bm{G_k}(\bm{\theta}, \bm{J})$ is the transformation matrix of joint $k$. The transformation from observation spaces to the canonical space, namely inverse LBS, takes a similar formulation with inverted transformation matrices.

\begin{figure}
    \begin{center}
    \includegraphics[width=\linewidth]{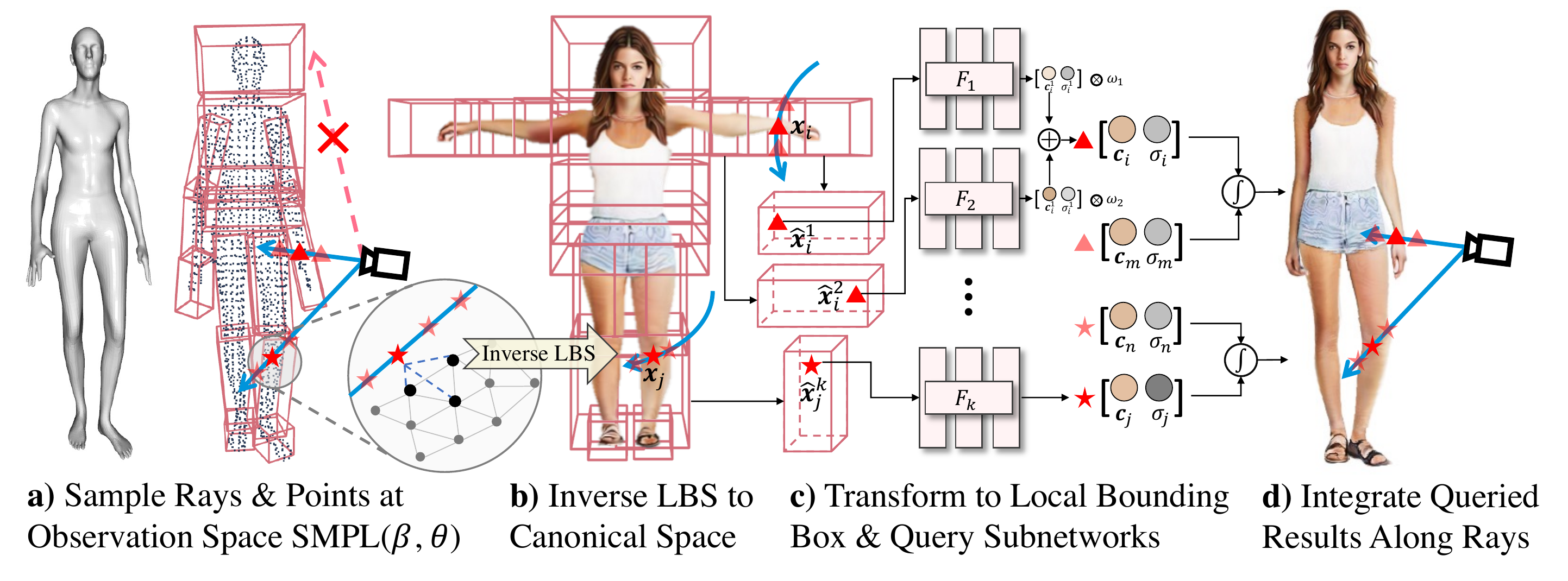}
    \end{center}
    \vspace{-12pt}
    \caption{\textbf{Rendering Process of the Compositional Human NeRF Representation.} For shape and pose specified by SMPL($\bm{\beta}$, $\bm{\theta}$), local bounding boxes are constructed. Rays that intersect with bounding boxes are sampled and transferred to the canonical space using inverse LBS. Subnetworks corresponding to bounding boxes are queried, results of which are integrated to produce final renderings.}
    \label{fig:comp_rep}
\end{figure}

\subsection{Compositional Human NeRF Representation} \label{method:repre}


The compositional human NeRF representation is defined as $\mathbb{F}_{\Phi}$, corresponding to a set of local bounding boxes $\mathbb{B}$. For each body part $k$, we use a subnetwork $F_k \in \mathbb{F}_{\Phi}$ to model the local bounding box $\{\bm{b}_{min}^{k}, \bm{b}_{max}^{k}\} \in \mathbb{B}$, as shown in Fig.~\ref{fig:comp_rep} b). For some point $\bm{x}_i$ in the canonical coordinate with direction $\bm{d}_i$ and falling inside the $k$-th bounding box, the corresponding radiance $\bm{c}_i^k$ and density $\sigma_i^k$ is queried by
\begin{equation} \label{eq:local_query}
    \{\bm{c}_i^k, \sigma_i^k\} = F_k(\hat{\bm{x}}_i^k, \bm{d}_i) \text{, where } \hat{\bm{x}}_i^k = \frac{2\bm{x}_i - (\bm{b}_{min}^{k} + \bm{b}_{max}^{k})}{\bm{b}_{max}^{k} - \bm{b}_{min}^{k}}.
\end{equation}
If the point $\bm{x}_i$ falls in multiple bounding boxes $\mathbb{A}_i$, a window function~\citep{lombardi2021mixture} is applied to linearly blend queried results. The blended radiance $\bm{c}_i$ and density $\sigma_i$ of $\bm{x}_i$ is calculated as
\begin{equation} \label{eq:windowfunction}
    \{\bm{c}_i, \sigma_i\} = \frac{1}{\sum\omega_a}\sum_{a\in\mathbb{A}_i}\omega_a\{\bm{c}_i^k, \sigma_i^k\} \text{, where } \omega_a = \text{exp}(-m(\hat{\bm{x}}_i^k(x)^n + \hat{\bm{x}}_i^k(y)^n + \hat{\bm{x}}_i^k(z)^n)).
\end{equation}
$m,n$ are chosen empirically. Different from \citet{palafox2022spams, su2022danbo}, we only query subnetworks whose bounding boxes contain query points. It increases the efficiency of the query process and saves computational resources.

Taking advantages of the compositional representation, we also adopt an efficient volume rendering algorithm. Previous methods need to sample points, query, and integrate for every pixel of the canvas, which wastes large amounts of computational resources on backgrounds. In contrast, for the compositional representation, we have pre-defined bounding boxes to filter useful rays, which is also the key for our method being able to train on high resolution.

As shown in Fig.~\ref{fig:comp_rep}, for the target pose $\bm{\theta}$, shape $\bm{\beta}$ and camera setup, our rendering algorithm $\mathcal{R}(\mathbb{F}_{\Phi}, \bm{\beta}, \bm{\theta}, \text{cam})$ is described as follows. Firstly, ray $\bm{r}(t)=\bm{o}+t\bm{d}$ is sampled for each pixel on the canvas. Then we transform the pre-defined bounding boxes $\mathbb{B}$ to the target pose $\bm{\theta}$ using transformation matrices $\bm{G_k}$ defined by SMPL. Rays that intersect with the transformed bounding boxes are kept for further rendering. Others are marked to be the background color. For ray $\bm{r}(t)=\bm{o}+t\bm{d}$ that intersects with single or multiple bounding boxes, we get the near and far bounds $t_n, t_f$. $N$ points are randomly sampled on each ray as
\begin{equation} \label{eq:samplepoints}
    t_i \sim \mathcal{U} \left[ t_n + \frac{i-1}{N}(t_f - t_n), t_n + \frac{i}{N}(t_f - t_n) \right].
\end{equation}
Next, we transform sampled points back to the canonical space using inverse LBS. Similar to \citet{zheng2021pamir}, we inverse not only the pose transformation, but also the shape/ pose blend shapes $\bm{B_S}(\bm{\beta}), \bm{B_P}(\bm{\theta})$ to be able to generalize to different body shapes. For sampled point $\bm{r}(t_i)$, the nearest $k$ points $\mathbb{N}=\{\bm{v}_1...\bm{v}_k\}$ are found among the vertices of the posed SMPL mesh $M(\bm{\beta}, \bm{\theta})$. The transformation of point $\bm{r}(t_i)$ from the observation space to the canonical space is defined as
\begin{equation} \label{eq:inverselbswhole}
    \begin{bmatrix}\bm{x}_i^0\\\bm{1}\end{bmatrix} = \sum_{v_j\in\mathbb{N}} \frac{\omega_j}{\sum\omega_j}(\bm{M}_j)^{-1} \begin{bmatrix}\bm{r}(t_i)\\\bm{1}\end{bmatrix} \text{, where } \bm{M}_j = \left( \sum_{k=1}^{K}w_k^j\bm{G}_k \right)
    \begin{bmatrix}\bm{I} & \bm{B_S^j} + \bm{B_P^j} \\ \bm{0} & \bm{I}\end{bmatrix}.
\end{equation}
$\omega_j = 1/\lVert \bm{r}(t_i) - \bm{v_j} \rVert$ is the inverse distance weight. $\bm{M}_j$ is the transformation matrix of the SMPL vertex $v_j$.
Then we query the compositional human NeRF representation $\mathbb{F}$ with point $\bm{x}_i^0$ to get its corresponding radiance $\bm{c}_i$ and density $\sigma_i$ as defined in Eq.~\ref{eq:local_query} and \ref{eq:windowfunction}. Finally, we integrate the queried results for the RGB value of ray $\bm{r}(t)$, as defined in Eq.~\ref{eq:nerf_dis}.

\begin{figure}
    \begin{center}
    \includegraphics[width=\linewidth]{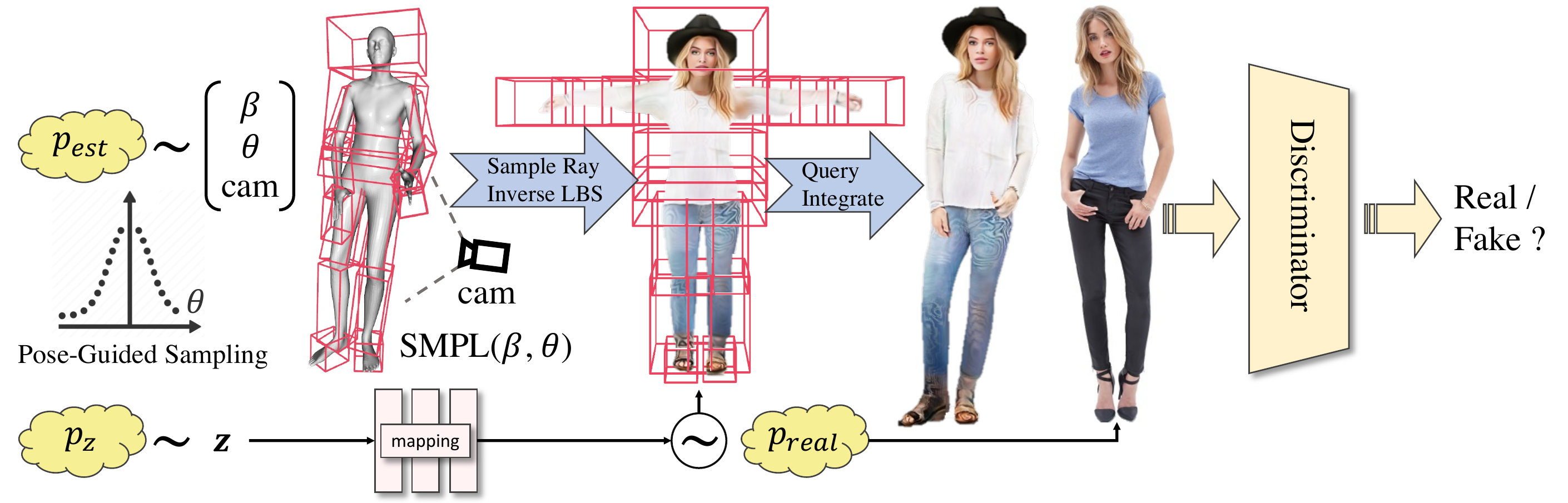}
    \end{center}
    \vspace{-12pt}
    \caption{\textbf{3D Human GAN Framework.} With the estimated SMPL and camera parameters distribution $p_{est}$, 3D humans are randomly sampled and rendered conditioned on $z\sim p_{z}$. The renderings are used for adversarial training against real 2D human image collections $p_{real}$.}
    \label{fig:gan}
\end{figure}

\subsection{3D Human GAN Framework} \label{method:ganframe}

With the compositional human NeRF representation, we construct a 3D human GAN framework as shown in Fig.~\ref{fig:gan}. The generator is defined as $G(\bm{z}, \bm{\beta}, \bm{\theta}, \text{cam}; \Phi_G) = \mathcal{R}(\mathbb{F}_{\Phi_G}(\bm{z}), \bm{\beta}, \bm{\theta}, \text{cam})$. Similar to pi-GAN~\citep{pigan}, each subnetwork of $\mathbb{F}_{\Phi}$ consists of stacked MLPs with SIREN activation~\citep{sitzmann2020implicit}. To generate fake samples, $\bm{z}\sim p_{\bm{z}}$ is sample from normal distribution. $\{\bm{\beta}, \bm{\theta}, \text{cam}\}\sim p_{est}$ are sampled from the estimated distribution from 2D image collections.
We use off-the-shelf tools~\citep{pavlakos2019expressive, kocabas2020vibe} to estimate $\{\bm{\beta}, \bm{\theta}, \text{cam}\}$ for the 2D image collections.
Unlike ENARF-GAN\citep{enarfgan}, where these variables are sampled from the distribution of motion datasets~\citep{mahmood2019amass}, the real 2D image collections do not necessarily share the similar pose distribution as that of motion datasets, especially for fashion datasets, \eg DeepFashion, where the pose distribution is imbalanced.
Finally, the fake samples $\bm{I}_f = G(\bm{z}, \bm{\beta}, \bm{\theta}, \text{cam}; \Phi_G)$, along with real samples $\bm{I}_r\sim p_{real}$ are sent to discriminator $D(I; \Phi_D)$ for adversarial training. For more implementation details, please refer to the supplementary material.

\subsection{Training} \label{method:training}

\noindent\textbf{Delta SDF Prediction.} Real-world 2D human image collections, especially fashion datasets, usually have imbalanced pose distribution. For example, as shown in Fig.~\ref{fig:distribution}, we plot the distribution of facing angles of DeepFashion. Such heavily imbalanced pose distribution makes it hard for the network to learn correct 3D information in an unsupervised way. Therefore, we propose to introduce strong human prior by utilizing the SMPL template geometry $\bm{d}_T(\bm{x})$ as the foundation of our human representation. Instead of directly predicting the SDF value $\bm{d}(\bm{x})$, we predict an SDF offset $\Delta\bm{d}(\bm{x})$ from the template~\citep{yifan2022geometryconsistent}. Then $\bm{d}_T(\bm{x}) + \Delta\bm{d}(\bm{x})$ is used as the actual SDF value of point $\bm{x}$.

\noindent\textbf{Pose-guided Sampling.} To facilitate effective 3D information learning from sparse 2D image collections, other than introducing a 3D human template, we propose to balance the input 2D images based on human poses. The intuition behind the pose-guided sampling is that different viewing angles should be sampled more evenly to allow effective learning of geometry. Empirically, among all human joints, we use the angle of the head to guide the sampling. Moreover, facial areas contain more information than other parts of the head. Front-view angles should be sampled more than other angles. Therefore, we choose to use a Gaussian distribution centered at the front-view angle $\mu_{\theta}$, with a standard deviation of $\sigma_{\theta}$. Specifically, $M$ bins are divided on the circle. For an image with the head angle falling in bin $m$, its probability $p_m$ of being sampled is defined as
\begin{equation} \label{eq:gaussiansample}
    p_m = \frac{1}{\sigma_{\theta}\sqrt{2\pi}}\text{exp}\left(-\frac{1}{2}\left( \frac{\theta_m - \mu_{\theta}}{\sigma_{\theta}} \right)^2 \right) \text{, where } \theta_m = \frac{2\pi m}{M}.
\end{equation}
We visualize the balanced distribution in Fig.~\ref{fig:distribution}. The network now has higher chances of seeing more side-views of human bodies, which helps better geometry generation.

\noindent\textbf{Loss Functions.} For the adversarial training, we use the non-saturating GAN loss with R1 regularization~\citep{mescheder2018training}, which is defined as
\begin{align} \label{eq:ganloss}
    \mathcal{L}_{\text{adv}}(\Phi_G, \Phi_D) &= \bm{E}_{\bm{z}\sim p_z, \{\bm{\beta}, \bm{\theta}, \text{cam}\}\sim p_{est}} [f(D(G(\bm{z}, \bm{\beta}, \bm{\theta}, \text{cam}; \Phi_G); \Phi_D))] \\
    &+ \bm{E}_{\bm{I}_r\sim p_{real}}[f(-D(\bm{I}_r; \Phi_D)) + \lambda|\nabla D(\bm{I}_r; \Phi_D)|^2],
\end{align}
where $f(u) = -\text{log}(1+\text{exp}(-u))$. Other than the adversarial loss, some regularization terms are introduced for the delta SDF prediction. Firstly, we want minimum offset from the template mesh to maintain plausible human shape, which gives the minimum offset loss $\mathcal{L}_{\text{off}} = \bm{E}_{\bm{x}}[\|\Delta d(\bm{x})\|_2^2]$. Secondly, to ensure that the predicted SDF values are physically valid~\citep{gropp2020implicit}, we penalize the derivation of delta SDF predictions to zero $\mathcal{L}_{\text{eik}} = \bm{E}_{\bm{x}}[\|\nabla (\Delta d(\bm{x})) \|_2^2]$. The overall loss is defined as $\mathcal{L} = \mathcal{L}_{\text{adv}} + \lambda_{\text{off}}\mathcal{L}_{\text{off}} + \lambda_{\text{eik}}\mathcal{L}_{\text{eik}}$, where $\lambda_{*}$ are loss weights defined empirically.

\begin{figure}
    \begin{center}
    \includegraphics[width=\linewidth]{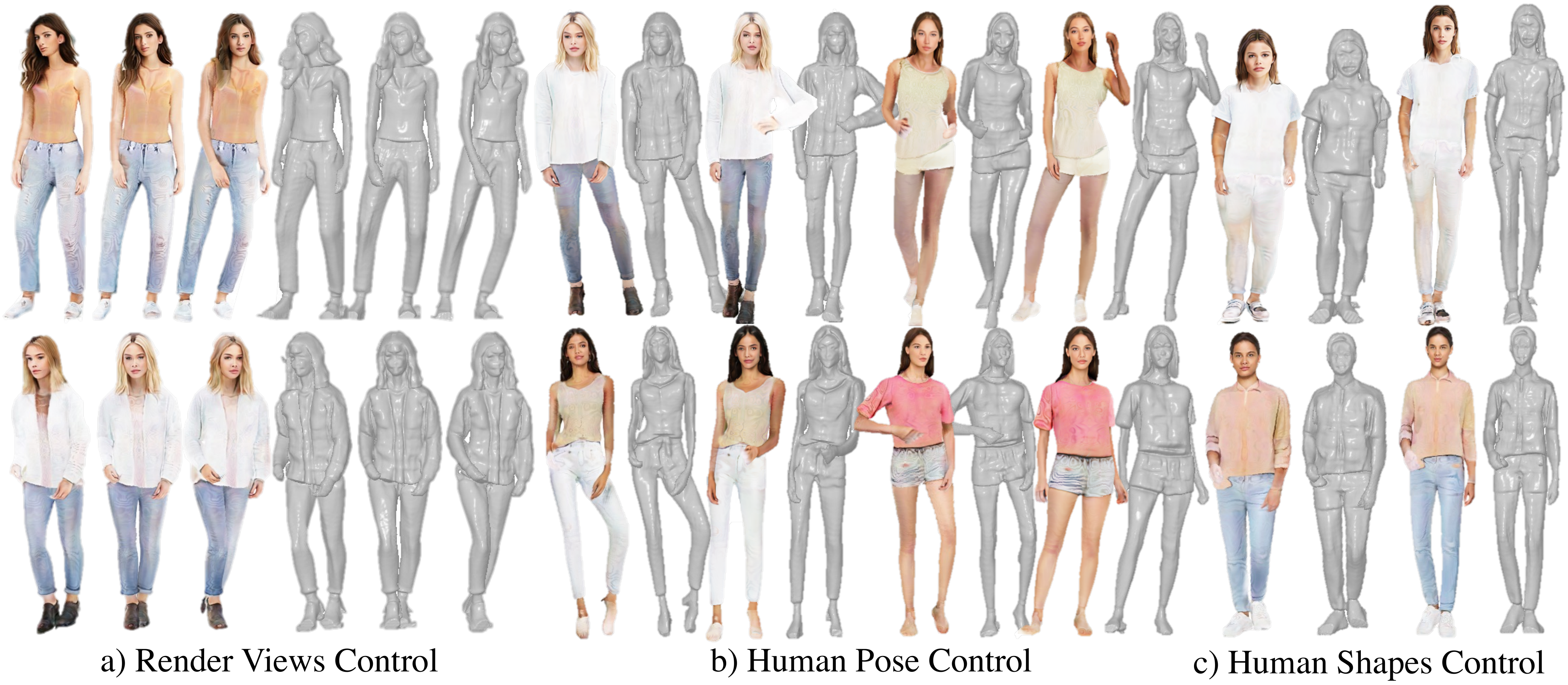}
    \end{center}
    \vspace{-12pt}
    \caption{\textbf{Generation Results of \nickname{}.} The 3D-aware nature and inherent human prior of \nickname{} enable explicit control over rendering views, human poses, and shapes.}
    \label{fig:novelview}
\end{figure}

\section{Experiments}

\subsection{Experimental Setup}

\noindent\textbf{Datasets.} We conduct experiments on four datasets: DeepFashion~\citep{deepfashion}, SHHQ~\citep{styleganhuman}, UBCFashion~\citep{ubcfashion} and AIST~\citep{aist}. The first two are sparse 2D image collections, meaning that each image has different identities and poses are sparse, which makes them more challenging. The last two are human video datasets containing different poses/ views of the same identities, which is easier for the task but lacks diversity. 

\noindent\textbf{Comparison Methods.} We mainly compare with three baselines. ENARF-GAN~\citep{enarfgan} makes the first attempt at human NeRF generation from 2D image collections. EG3D~\citep{eg3d} and StyleSDF~\citep{stylesdf} are state-of-the-art methods for 3D-aware generation, both requiring super-resolution modules to achieve high-resolution generation.

\noindent\textbf{Evaluation Metrics.} To evaluate the quality of rendered images, we adopt Frechet Inception Distance (FID)~\citep{heusel2017gans} and Kernel Inception Distance (KID)~\citep{binkowski2018demystifying}. Following ENARF-GAN, we use Percentage of Correct Keypoints (PCKh@0.5)~\citep{andriluka20142d} to evaluate the correctness of generated poses. Note that PCKh@0.5 can only be calculated on methods that can control generated poses, \ie ENARF-GAN and \nickname{}. To evaluate the correctness of geometry, we use an off-the-shelf tool~\citep{Ranftl2022} to estimate depth from the generated images and compare it with generated depths.
$50K$ samples padded to square are used to compute FID and KID. PCKh@0.5 and Depth are evaluated on $5K$ samples.

\begin{figure}
    \begin{center}
    \includegraphics[width=\linewidth]{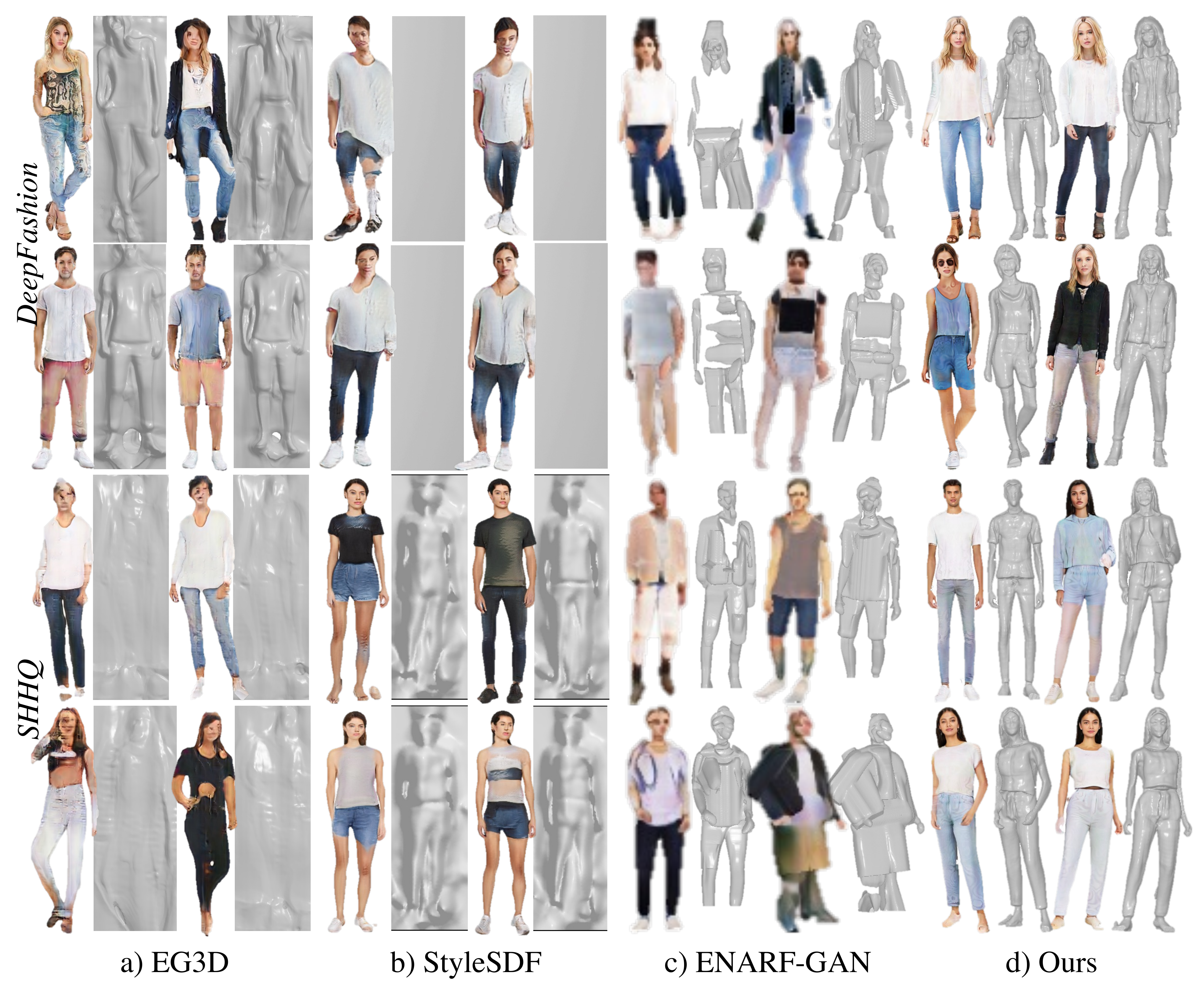}
    \end{center}
    \vspace{-12pt}
    \caption{\textbf{Qualitative Comparison Between \nickname{} and Baseline Methods.} Rendered 2D images and corresponding meshes are placed side-by-side. Both the 2D renderings and 3D meshes generated by our method achieve the best quality among SOTA methods. Zoom in for the best view.}
    \label{fig:main_comp}
    \vspace{-10pt}
\end{figure}


\subsection{Qualitative Evaluations}

\noindent\textbf{Generation Results and Controlling Ability of \nickname{}.}
As shown in Fig.~\ref{fig:novelview} a), \nickname{} is capable of generating high-quality renderings in novel views and remain multi-view consistency. Due to the inherent human prior in our model design, \nickname{} can control poses and shapes of the generated 3D human by changing $\bm{\beta}$ and $\bm{\theta}$ of SMPL. We show novel pose and shape generation results in Fig.~\ref{fig:novelview} b)\& c). We refer readers to the supplementary PDF and video for more qualitative results.

\noindent\textbf{Comparison with Baseline Methods.}
We show the renderings and corresponding meshes generated by baselines and our method in Fig.~\ref{fig:main_comp}. EG3D trained on DeepFashion, as well as StyleSDF trained on SHHQ, generate reasonable RGB renderings and geometry. However, without explicit human modeling, complex human poses make it hard to align and model 3D humans in observation spaces, which leads to distorted generation. Moreover, because of the use of super resolution, their geometry is only trained under low resolution ($64^2$) and therefore lacks details. EG3D trained on SHHQ and StyleSDF trained on DeepFashion fail to capture 3D information and collapse to the trivial solution of painting on billboards. Limited by the inefficient representation and computational resources, ENARF-GAN can only be trained at a resolution of $128^2$, which leads to low-quality rendering results. Besides, lacking human prior makes ENARF-GAN hard to capture correct 3D information of human from sparse 2D image collections, which results in broken meshes. \nickname{}, in contrast, generates high-quality human renderings on both datasets. We also succeeded in learning reasonable 3D human geometry from 2D image collections with sparse viewing angles and poses, thanks to the strong human prior and the pose-guided sampling strategy. Due to space limitations, we only show results of DeepFashion and SHHQ here. For visual comparisons on UBCFashion and AIST, please refer to the supplementary material.

\begin{table}[t]
\caption{Comparison with State-of-the-Art Methods. * The training code of ENARF-GAN is implemented based on the official inference code.}
\vspace{-12pt}
\label{tab:main}
\begin{center}
\begin{tabular}{l|cccc|cccc}
    \Xhline{1pt}
    \multirow{2}{*}{Methods, Resolution} & \multicolumn{4}{c|}{\textit{DeepFashion}} & \multicolumn{4}{c}{\textit{SHHQ}} \\
    & FID$\downarrow$ & KID$\downarrow$ & PCK$\uparrow$ & Depth$\downarrow$ & FID$\downarrow$ & KID$\downarrow$ & PCK$\uparrow$ & Depth$\downarrow$ \\
    \hline\hline
    EG3D, $512^2$ & 26.38 & 0.014 & - & 0.0779 & 32.96 & 0.033 & - & 0.0296  \\
    StyleSDF, $512^2$ & 92.40 & 0.136 & - & 0.0359 & 14.12 & 0.010 & - & 0.0300 \\
    ENARF-GAN*, $128^2$ & 77.03 & 0.114 & 43.74 & 0.1151 & 80.54 & 0.102 & 40.17 & 0.1241 \\
    Ours, $512^2$ & \textbf{15.91} & \textbf{0.011} & \textbf{87.50} & \textbf{0.0272} & \textbf{11.99} & \textbf{0.009} & \textbf{88.95} & \textbf{0.0177} \\
    \Xhline{1pt}
    \multirow{2}{*}{Methods, Resolution} & \multicolumn{4}{c|}{\textit{UBCFashion}} & \multicolumn{4}{c}{\textit{AIST}} \\
    & FID$\downarrow$ & KID$\downarrow$ & PCK$\uparrow$ & Depth$\downarrow$ & FID$\downarrow$ & KID$\downarrow$ & PCK$\uparrow$ & Depth$\downarrow$ \\
    \hline\hline
    EG3D, $512^2$ & 23.95 & \textbf{0.009} & - & 0.1163 & 34.76 & 0.022 & - & 0.1165 \\
    StyleSDF, $512^2$ & 18.52 & 0.011 & - & 0.0311 & 199.5 & 0.225 & - & 0.0236 \\
    ENARF-GAN*, $128^2$ & - & - & - & - & 73.07 & 0.075 & 42.85 & 0.1128 \\
    Ours, $512^2$ & \textbf{12.61} & 0.010 & \textbf{99.17} & \textbf{0.0090} & \textbf{19.40} & \textbf{0.010} & \textbf{83.15} & \textbf{0.0126} \\
    \Xhline{1pt}
\end{tabular}
\vspace{-6pt}
\end{center}
\end{table}

\subsection{Quantitative Evaluations}
As shown in Tab.~\ref{tab:main}, our method leads all metrics in four datasets. \nickname{} outperforms ENARF-GAN in all settings thanks to our high-resolution training ability. EG3D and StyleSDF, as the SOTA methods in the 3D generation, can achieve reasonable scores in some settings (\eg StyleSDF achieves 18.52 FID on UBCFashion) for their super-resolution modules. But they also fail on some datasets (\eg StyleSDF fails on AIST with 199.5 FID) for complexity in human poses. In the contrast, \nickname{} achieves the best FID/KID scores under all settings. Moreover, unlike EG3D or StyleSDF, \nickname{} can control the generated pose and achieve higher PCKh@0.5 score than ENARF-GAN. For the geometry part, we also achieve the lowest depth error, which shows the importance of natively high-resolution training.

\begin{table}[t]
\centering
\makebox[0pt][c]{\parbox{1.0\textwidth}{
\begin{minipage}[c]{0.34\hsize} 
    \centering
    \caption{Results of Ablation Study. $^\dagger$Depth is evaluated against SMPL depth. We report $^\dagger$Depth$\times10^3$ for simplicity.}
    \label{tab:ablation}
    \vspace{-2pt}
    \small{
    \addtolength{\tabcolsep}{-3.5pt}
    \begin{tabular}{l|ccc}
    \Xhline{1pt}
    Methods & FID$\downarrow$ & $^\dagger$Depth$\downarrow$ \\
    \hline\hline
    Baseline, \scriptsize{$256^2$} & 31.14 & 3.57 \\
    + Composite, \scriptsize{$512^2$} & 17.81 & 5.02 \\
    + Delta SDF, \scriptsize{$512^2$} & \textbf{15.62} & 3.69 \\
    + Pose-guide, \scriptsize{$512^2$} & 15.91 & \textbf{3.04} \\
    \Xhline{1pt}
    \end{tabular}
    }
\end{minipage}
\hfill
\begin{minipage}[c]{0.28\hsize} 
    \centering
    \caption{Trade-Off Between RGB and Geometry.}
    \label{tab:gaussian}
    \small{
    \addtolength{\tabcolsep}{-3pt}
    \begin{tabular}{l|ccc}
    \Xhline{1pt}
    Distribution & FID$\downarrow$ & $^\dagger$Depth$\downarrow$ \\
    \hline\hline
    Original & 15.62 & 3.69 \\
    $\sigma_\theta=15^{\circ}$ & 15.91 & 3.04 \\
    $\sigma_\theta=30^{\circ}$ & 19.05 & 2.58 \\
    $\sigma_\theta=45^{\circ}$ & 19.56 & 2.65 \\
    $\sigma_\theta=60^{\circ}$ & 25.08 & 2.91 \\
    Uniform & 25.82 & 2.92 \\
    \Xhline{1pt}
    \end{tabular}
    }
\end{minipage}
\hfill
\begin{minipage}[c]{0.33\hsize} 
    \begin{center}
    \includegraphics[width=0.95\linewidth]{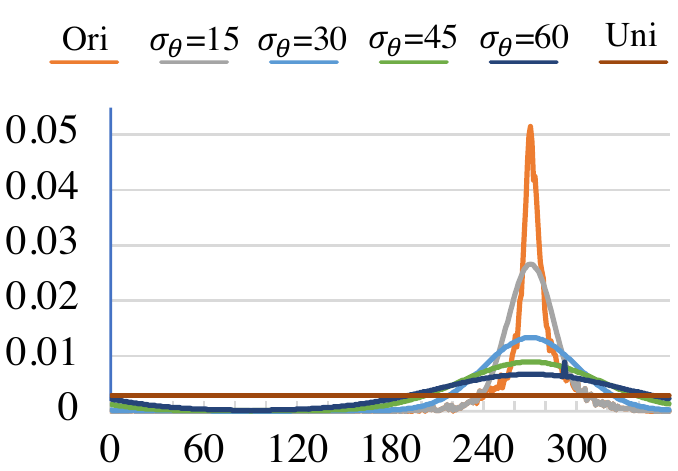}
    \end{center}
    \vspace{-12pt}
    \captionof{figure}{PDF of Different Pose-Guided Sampling Distributions.}
    \label{fig:distribution}
\end{minipage}
}}
\vspace{-6pt}
\end{table}

\subsection{Ablation Studies}

\noindent\textbf{Ablation on Method Designs.}
To validate the effectiveness of our designs on \nickname{}, we subsequently add different designs on a baseline method, which uses one large network to model the canonical space. Experiments are conducted on DeepFashion. The results are reported in Tab.~\ref{tab:ablation}. Limited by the inefficient representation, the baseline (``Baseline'') can only be trained at $256\times 128$, which results in the worst FID score. Adding compositional design (``+Composite'') makes the network efficient enough to be trained at a higher resolution of $512\times 256$ and achieve higher generation quality. We further introduce human prior by predicting delta SDF (``+Delta SDF''), which gives the best FID score and lower depth error. Finally, using the pose-guided sampling (``+Pose-guide''), we further decrease the depth error, which means better geometry. However, the FID score slightly increases, which is discussed in the next paragraph. We refer readers to the supplementary material for qualitative evaluations of ablation studies.

\noindent\textbf{Analysis on Pose-Guided Sampling.}
We further analyze the importance of the sampling strategy in 3D human GAN training. Three types of distributions $p_{est}$ are experimented, including the original dataset distribution (``Original''), pose-guided Gaussian distribution (``$\sigma_\theta=*$''), and pose-guided uniform distribution (``Uniform''). The results are reported in Tab.~\ref{tab:gaussian}. Firstly, uniform sampling is not a good strategy, as shown by its high FID score. This is because the information density is different between different parts of human. Faces require more training iterations. Secondly, the original distribution gives the best visual quality but the worst geometry. As shown in Fig.~\ref{fig:distribution}, the original distribution leads to the network mostly being trained on front-view images. It could result in the trivial solution of painting on billboards. Though using delta SDF prediction can alleviate the problem to some extent, the geometry is still not good enough. Thirdly, the pose-guided Gaussian sampling can avoid damaging visual quality too much and improve geometry learning. As the standard deviation $\sigma_{\theta}$ increases, FID increases while the depth error decreases. Therefore, it is a trade-off between visual quality and geometry quality. In our final experiments, we choose $\sigma_\theta=15^{\circ}$ which is a satisfying balance between the two factors.

\begin{figure}
    \begin{center}
    \includegraphics[width=\linewidth]{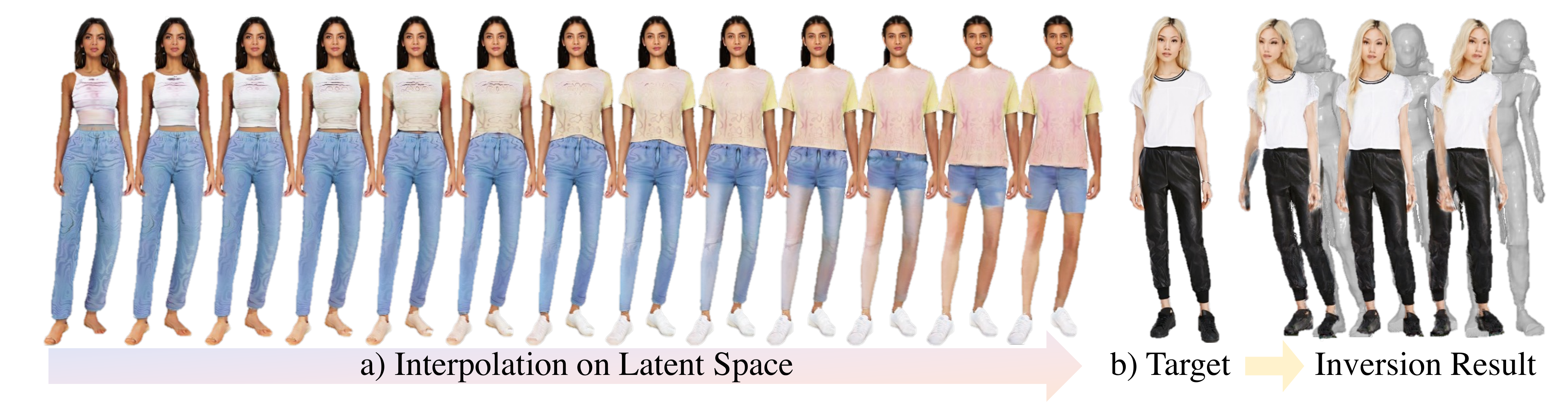}
    \end{center}
    \vspace{-15pt}
    \caption{\textbf{Applications of \nickname{}.} a) Interpolation on the latent space gives smooth transition between two samples. b) Inversion results (right) of the target image (left).}
    \label{fig:application}
    \vspace{-10pt}
\end{figure}

\subsection{Applications}

\noindent\textbf{Interpolation on Latent Space.}
As shown in Fig.~\ref{fig:application} a), we linearly interpolate two latent codes to generate a smooth transition between them, showing that the latent space learned by \nickname{} is semantically meaningful. More results are provided in the supplementary video.

\noindent\textbf{Inversion.}
We use Pivotal Tuning Inversion (PTI)~\citep{roich2021pivotal}  to inverse the target image and show the results in Fig.~\ref{fig:application} b). Reasonable novel view synthesis results can be achieved. The geometry, however, fails to capture geometry details corresponding to RGB renderings, which can be caused by the second stage generator fine-tuning of PTI. Nevertheless, we demonstrate the potential of \nickname{} in more related downstream tasks.

\section{Discussion}
To conclude, we propose a high-quality unconditional 3D human generation model \nickname{} that only requires 2D image collections for training. 
We design a compositional human NeRF representation for efficient GAN training. To train on the challenging 2D image collections with sparse viewing angles and human poses, \eg DeepFashion, strong human prior and pose-guided sampling are introduced for better GAN learning. On four large-scale 2D human datasets, we achieve state-of-the-art generation results at a high resolution of $512\times256$.

\noindent\textbf{Limitations:} 
\textbf{1)} There still exists visible circular artifacts in the renderings, which might be caused by the SIREN activation. A better base representation, \eg tri-plane of EG3D, and a 2D decoder might solve the issue.
\textbf{2)} The estimation of SMPL parameters from 2D image collections is not accurate, which leads to a distribution shift from the real pose distribution and possibly compromises generation results. Refining SMPL estimation during training would make a good future work.
\textbf{3)} Limited by our tight 3D human representation, it is hard to model loose clothes like dresses. Using separate modules to handle loose clothes might be a promising direction.

\subsubsection*{Ethics Statement}
Although the results of \nickname{} are yet to the point where they can fake human eyes, we still need to be aware of its potential ethical issues. The generated 3D humans might be misused to create contents that are misleading. \nickname{} can also be used to invert real human images, which can be used to create fake videos of real humans and cause negative social impacts. Moreover, the generated 3D humans might be biased, which is caused by the inherent distribution of training datasets. We make our best effort to demonstrate the impartiality of \nickname{} in Fig.~\ref{fig:teaser}.

\subsubsection*{Reproducibility Statement}
Our method is thoroughly described in Sec.~\ref{method}. Together with implementation details included in the supplementary material, the reproducibility is ensured. Moreover, our code will be released upon acceptance.


\subsubsection*{Acknowledgments}
This work is supported by NTU NAP, MOE AcRF Tier 2 (T2EP20221-0033), and under the RIE2020 Industry Alignment Fund – Industry Collaboration Projects (IAF-ICP) Funding Initiative, as well as cash and in-kind contribution from the industry partner(s).

\bibliography{iclr2023_conference}

\begin{thebibliography}{73}
\providecommand{\natexlab}[1]{#1}
\providecommand{\url}[1]{\texttt{#1}}
\expandafter\ifx\csname urlstyle\endcsname\relax
  \providecommand{\doi}[1]{doi: #1}\else
  \providecommand{\doi}{doi: \begingroup \urlstyle{rm}\Url}\fi

\bibitem[Andriluka et~al.(2014)Andriluka, Pishchulin, Gehler, and
  Schiele]{andriluka20142d}
Mykhaylo Andriluka, Leonid Pishchulin, Peter Gehler, and Bernt Schiele.
\newblock 2d human pose estimation: New benchmark and state of the art
  analysis.
\newblock In \emph{Proceedings of the IEEE Conference on computer Vision and
  Pattern Recognition}, pp.\  3686--3693, 2014.

\bibitem[Bergman et~al.(2022)Bergman, Kellnhofer, Wang, Chan, Lindell, and
  Wetzstein]{eg3dhuman}
Alexander~W Bergman, Petr Kellnhofer, Yifan Wang, Eric~R Chan, David~B Lindell,
  and Gordon Wetzstein.
\newblock Generative neural articulated radiance fields.
\newblock \emph{arXiv preprint arXiv:2206.14314}, 2022.

\bibitem[Bi{\'n}kowski et~al.(2018)Bi{\'n}kowski, Sutherland, Arbel, and
  Gretton]{binkowski2018demystifying}
Miko{\l}aj Bi{\'n}kowski, Danica~J Sutherland, Michael Arbel, and Arthur
  Gretton.
\newblock Demystifying mmd gans.
\newblock \emph{arXiv preprint arXiv:1801.01401}, 2018.

\bibitem[Cai et~al.(2022)Cai, Ren, Zeng, Lin, Yu, Wang, Fan, Gao, Yu, Pan,
  et~al.]{humman}
Zhongang Cai, Daxuan Ren, Ailing Zeng, Zhengyu Lin, Tao Yu, Wenjia Wang,
  Xiangyu Fan, Yang Gao, Yifan Yu, Liang Pan, et~al.
\newblock Humman: Multi-modal 4d human dataset for versatile sensing and
  modeling.
\newblock \emph{arXiv preprint arXiv:2204.13686}, 2022.

\bibitem[Chan et~al.(2021)Chan, Monteiro, Kellnhofer, Wu, and Wetzstein]{pigan}
Eric~R Chan, Marco Monteiro, Petr Kellnhofer, Jiajun Wu, and Gordon Wetzstein.
\newblock pi-gan: Periodic implicit generative adversarial networks for
  3d-aware image synthesis.
\newblock In \emph{Proceedings of the IEEE/CVF conference on computer vision
  and pattern recognition}, pp.\  5799--5809, 2021.

\bibitem[Chan et~al.(2022)Chan, Lin, Chan, Nagano, Pan, De~Mello, Gallo,
  Guibas, Tremblay, Khamis, et~al.]{eg3d}
Eric~R Chan, Connor~Z Lin, Matthew~A Chan, Koki Nagano, Boxiao Pan, Shalini
  De~Mello, Orazio Gallo, Leonidas~J Guibas, Jonathan Tremblay, Sameh Khamis,
  et~al.
\newblock Efficient geometry-aware 3d generative adversarial networks.
\newblock In \emph{Proceedings of the IEEE/CVF Conference on Computer Vision
  and Pattern Recognition}, pp.\  16123--16133, 2022.

\bibitem[Chen et~al.(2021)Chen, Zhang, Kang, Zhe, Bao, Jia, and
  Lu]{chen2021animatable}
Jianchuan Chen, Ying Zhang, Di~Kang, Xuefei Zhe, Linchao Bao, Xu~Jia, and
  Huchuan Lu.
\newblock Animatable neural radiance fields from monocular rgb videos.
\newblock \emph{arXiv preprint arXiv:2106.13629}, 2021.

\bibitem[Chen et~al.(2022)Chen, Jiang, Song, Yang, Black, Geiger, and
  Hilliges]{chen2022gdna}
Xu~Chen, Tianjian Jiang, Jie Song, Jinlong Yang, Michael~J Black, Andreas
  Geiger, and Otmar Hilliges.
\newblock gdna: Towards generative detailed neural avatars.
\newblock \emph{arXiv}, 2022.

\bibitem[Deng et~al.(2022)Deng, Yang, Xiang, and Tong]{GRAM}
Yu~Deng, Jiaolong Yang, Jianfeng Xiang, and Xin Tong.
\newblock Gram: Generative radiance manifolds for 3d-aware image generation.
\newblock In \emph{Proceedings of the IEEE/CVF Conference on Computer Vision
  and Pattern Recognition (CVPR)}, pp.\  10673--10683, June 2022.

\bibitem[Fr{\"u}hst{\"u}ck et~al.(2022)Fr{\"u}hst{\"u}ck, Singh, Shechtman,
  Mitra, Wonka, and Lu]{fruhstuck2022insetgan}
Anna Fr{\"u}hst{\"u}ck, Krishna~Kumar Singh, Eli Shechtman, Niloy~J Mitra,
  Peter Wonka, and Jingwan Lu.
\newblock Insetgan for full-body image generation.
\newblock In \emph{Proceedings of the IEEE/CVF Conference on Computer Vision
  and Pattern Recognition}, pp.\  7723--7732, 2022.

\bibitem[Fu et~al.(2022)Fu, Li, Jiang, Lin, Qian, Loy, Wu, and
  Liu]{styleganhuman}
Jianglin Fu, Shikai Li, Yuming Jiang, Kwan-Yee Lin, Chen Qian, Chen~Change Loy,
  Wayne Wu, and Ziwei Liu.
\newblock Stylegan-human: A data-centric odyssey of human generation.
\newblock \emph{arXiv preprint arXiv:2204.11823}, 2022.

\bibitem[Goodfellow et~al.(2020)Goodfellow, Pouget-Abadie, Mirza, Xu,
  Warde-Farley, Ozair, Courville, and Bengio]{goodfellow2020generative}
Ian Goodfellow, Jean Pouget-Abadie, Mehdi Mirza, Bing Xu, David Warde-Farley,
  Sherjil Ozair, Aaron Courville, and Yoshua Bengio.
\newblock Generative adversarial networks.
\newblock \emph{Communications of the ACM}, 63\penalty0 (11):\penalty0
  139--144, 2020.

\bibitem[Grigorev et~al.(2021)Grigorev, Iskakov, Ianina, Bashirov, Zakharkin,
  Vakhitov, and Lempitsky]{stylepeople}
Artur Grigorev, Karim Iskakov, Anastasia Ianina, Renat Bashirov, Ilya
  Zakharkin, Alexander Vakhitov, and Victor Lempitsky.
\newblock Stylepeople: A generative model of fullbody human avatars.
\newblock In \emph{Proceedings of the IEEE/CVF Conference on Computer Vision
  and Pattern Recognition}, pp.\  5151--5160, 2021.

\bibitem[Gropp et~al.(2020)Gropp, Yariv, Haim, Atzmon, and
  Lipman]{gropp2020implicit}
Amos Gropp, Lior Yariv, Niv Haim, Matan Atzmon, and Yaron Lipman.
\newblock Implicit geometric regularization for learning shapes.
\newblock \emph{arXiv preprint arXiv:2002.10099}, 2020.

\bibitem[Gu et~al.(2021)Gu, Liu, Wang, and Theobalt]{stylenerf}
Jiatao Gu, Lingjie Liu, Peng Wang, and Christian Theobalt.
\newblock Stylenerf: A style-based 3d-aware generator for high-resolution image
  synthesis.
\newblock \emph{arXiv preprint arXiv:2110.08985}, 2021.

\bibitem[Habermann et~al.(2021)Habermann, Liu, Xu, Zollhoefer, Pons-Moll, and
  Theobalt]{habermann2021real}
Marc Habermann, Lingjie Liu, Weipeng Xu, Michael Zollhoefer, Gerard Pons-Moll,
  and Christian Theobalt.
\newblock Real-time deep dynamic characters.
\newblock \emph{ACM Transactions on Graphics (TOG)}, 40\penalty0 (4):\penalty0
  1--16, 2021.

\bibitem[Henzler et~al.(2019)Henzler, Mitra, and Ritschel]{platogan}
Philipp Henzler, Niloy~J Mitra, and Tobias Ritschel.
\newblock Escaping plato's cave: 3d shape from adversarial rendering.
\newblock In \emph{Proceedings of the IEEE/CVF International Conference on
  Computer Vision}, pp.\  9984--9993, 2019.

\bibitem[Heusel et~al.(2017)Heusel, Ramsauer, Unterthiner, Nessler, and
  Hochreiter]{heusel2017gans}
Martin Heusel, Hubert Ramsauer, Thomas Unterthiner, Bernhard Nessler, and Sepp
  Hochreiter.
\newblock Gans trained by a two time-scale update rule converge to a local nash
  equilibrium.
\newblock \emph{Advances in neural information processing systems}, 30, 2017.

\bibitem[Hong et~al.(2021)Hong, Pan, Cai, and Liu]{garment4d}
Fangzhou Hong, Liang Pan, Zhongang Cai, and Ziwei Liu.
\newblock Garment4d: Garment reconstruction from point cloud sequences.
\newblock \emph{Advances in Neural Information Processing Systems},
  34:\penalty0 27940--27951, 2021.

\bibitem[Hong et~al.(2022{\natexlab{a}})Hong, Pan, Cai, and Liu]{hcmoco}
Fangzhou Hong, Liang Pan, Zhongang Cai, and Ziwei Liu.
\newblock Versatile multi-modal pre-training for human-centric perception.
\newblock In \emph{Proceedings of the IEEE/CVF Conference on Computer Vision
  and Pattern Recognition}, pp.\  16156--16166, 2022{\natexlab{a}}.

\bibitem[Hong et~al.(2022{\natexlab{b}})Hong, Zhang, Pan, Cai, Yang, and
  Liu]{avatarclip}
Fangzhou Hong, Mingyuan Zhang, Liang Pan, Zhongang Cai, Lei Yang, and Ziwei
  Liu.
\newblock Avatarclip: Zero-shot text-driven generation and animation of 3d
  avatars.
\newblock \emph{arXiv preprint arXiv:2205.08535}, 2022{\natexlab{b}}.

\bibitem[Jiang et~al.(2022{\natexlab{a}})Jiang, Hong, Bao, and
  Zhang]{jiang2022selfrecon}
Boyi Jiang, Yang Hong, Hujun Bao, and Juyong Zhang.
\newblock Selfrecon: Self reconstruction your digital avatar from monocular
  video.
\newblock In \emph{Proceedings of the IEEE/CVF Conference on Computer Vision
  and Pattern Recognition}, pp.\  5605--5615, 2022{\natexlab{a}}.

\bibitem[Jiang et~al.(2022{\natexlab{b}})Jiang, Yi, Samei, Tuzel, and
  Ranjan]{jiang2022neuman}
Wei Jiang, Kwang~Moo Yi, Golnoosh Samei, Oncel Tuzel, and Anurag Ranjan.
\newblock Neuman: Neural human radiance field from a single video.
\newblock \emph{arXiv preprint arXiv:2203.12575}, 2022{\natexlab{b}}.

\bibitem[Jiang et~al.(2022{\natexlab{c}})Jiang, Yang, Qiu, Wu, Loy, and
  Liu]{jiang2022text2human}
Yuming Jiang, Shuai Yang, Haonan Qiu, Wayne Wu, Chen~Change Loy, and Ziwei Liu.
\newblock Text2human: Text-driven controllable human image generation.
\newblock \emph{ACM Transactions on Graphics (TOG)}, 41\penalty0 (4):\penalty0
  1--11, 2022{\natexlab{c}}.
\newblock \doi{10.1145/3528223.3530104}.

\bibitem[Karras et~al.(2019)Karras, Laine, and Aila]{stylegan1}
Tero Karras, Samuli Laine, and Timo Aila.
\newblock A style-based generator architecture for generative adversarial
  networks.
\newblock In \emph{Proceedings of the IEEE/CVF conference on computer vision
  and pattern recognition}, pp.\  4401--4410, 2019.

\bibitem[Karras et~al.(2020)Karras, Laine, Aittala, Hellsten, Lehtinen, and
  Aila]{stylegan2}
Tero Karras, Samuli Laine, Miika Aittala, Janne Hellsten, Jaakko Lehtinen, and
  Timo Aila.
\newblock Analyzing and improving the image quality of stylegan.
\newblock In \emph{Proceedings of the IEEE/CVF conference on computer vision
  and pattern recognition}, pp.\  8110--8119, 2020.

\bibitem[Kingma \& Ba(2014)Kingma and Ba]{kingma2014adam}
Diederik~P Kingma and Jimmy Ba.
\newblock Adam: A method for stochastic optimization.
\newblock \emph{arXiv preprint arXiv:1412.6980}, 2014.

\bibitem[Kocabas et~al.(2020)Kocabas, Athanasiou, and Black]{kocabas2020vibe}
Muhammed Kocabas, Nikos Athanasiou, and Michael~J Black.
\newblock Vibe: Video inference for human body pose and shape estimation.
\newblock In \emph{Proceedings of the IEEE/CVF conference on computer vision
  and pattern recognition}, pp.\  5253--5263, 2020.

\bibitem[Kolotouros et~al.(2019)Kolotouros, Pavlakos, Black, and
  Daniilidis]{SPIN}
Nikos Kolotouros, Georgios Pavlakos, Michael~J Black, and Kostas Daniilidis.
\newblock Learning to reconstruct 3d human pose and shape via model-fitting in
  the loop.
\newblock In \emph{Proceedings of the IEEE/CVF International Conference on
  Computer Vision}, pp.\  2252--2261, 2019.

\bibitem[Lewis et~al.(2021)Lewis, Varadharajan, and
  Kemelmacher-Shlizerman]{tryongan}
Kathleen~M Lewis, Srivatsan Varadharajan, and Ira Kemelmacher-Shlizerman.
\newblock Tryongan: Body-aware try-on via layered interpolation.
\newblock \emph{ACM Transactions on Graphics (TOG)}, 40\penalty0 (4):\penalty0
  1--10, 2021.

\bibitem[Liu et~al.(2021)Liu, Habermann, Rudnev, Sarkar, Gu, and
  Theobalt]{neuralactor}
Lingjie Liu, Marc Habermann, Viktor Rudnev, Kripasindhu Sarkar, Jiatao Gu, and
  Christian Theobalt.
\newblock Neural actor: Neural free-view synthesis of human actors with pose
  control.
\newblock \emph{ACM Transactions on Graphics (TOG)}, 40\penalty0 (6):\penalty0
  1--16, 2021.

\bibitem[Liu et~al.(2016)Liu, Luo, Qiu, Wang, and Tang]{deepfashion}
Ziwei Liu, Ping Luo, Shi Qiu, Xiaogang Wang, and Xiaoou Tang.
\newblock Deepfashion: Powering robust clothes recognition and retrieval with
  rich annotations.
\newblock In \emph{Proceedings of IEEE Conference on Computer Vision and
  Pattern Recognition (CVPR)}, June 2016.

\bibitem[Lombardi et~al.(2021)Lombardi, Simon, Schwartz, Zollhoefer, Sheikh,
  and Saragih]{lombardi2021mixture}
Stephen Lombardi, Tomas Simon, Gabriel Schwartz, Michael Zollhoefer, Yaser
  Sheikh, and Jason Saragih.
\newblock Mixture of volumetric primitives for efficient neural rendering.
\newblock \emph{ACM Transactions on Graphics (TOG)}, 40\penalty0 (4):\penalty0
  1--13, 2021.

\bibitem[Loper et~al.(2015)Loper, Mahmood, Romero, Pons-Moll, and Black]{smpl}
Matthew Loper, Naureen Mahmood, Javier Romero, Gerard Pons-Moll, and Michael~J
  Black.
\newblock Smpl: A skinned multi-person linear model.
\newblock \emph{ACM transactions on graphics (TOG)}, 34\penalty0 (6):\penalty0
  1--16, 2015.

\bibitem[Mahmood et~al.(2019)Mahmood, Ghorbani, Troje, Pons-Moll, and
  Black]{mahmood2019amass}
Naureen Mahmood, Nima Ghorbani, Nikolaus~F Troje, Gerard Pons-Moll, and
  Michael~J Black.
\newblock Amass: Archive of motion capture as surface shapes.
\newblock In \emph{Proceedings of the IEEE/CVF international conference on
  computer vision}, pp.\  5442--5451, 2019.

\bibitem[Mescheder et~al.(2018)Mescheder, Geiger, and
  Nowozin]{mescheder2018training}
Lars Mescheder, Andreas Geiger, and Sebastian Nowozin.
\newblock Which training methods for gans do actually converge?
\newblock In \emph{International conference on machine learning}, pp.\
  3481--3490. PMLR, 2018.

\bibitem[Mihajlovic et~al.(2022)Mihajlovic, Saito, Bansal, Zollhoefer, and
  Tang]{mihajlovic2022coap}
Marko Mihajlovic, Shunsuke Saito, Aayush Bansal, Michael Zollhoefer, and Siyu
  Tang.
\newblock Coap: Compositional articulated occupancy of people.
\newblock In \emph{Proceedings of the IEEE/CVF Conference on Computer Vision
  and Pattern Recognition}, pp.\  13201--13210, 2022.

\bibitem[Mildenhall et~al.(2020)Mildenhall, Srinivasan, Tancik, Barron,
  Ramamoorthi, and Ng]{nerf}
Ben Mildenhall, Pratul~P Srinivasan, Matthew Tancik, Jonathan~T Barron, Ravi
  Ramamoorthi, and Ren Ng.
\newblock Nerf: Representing scenes as neural radiance fields for view
  synthesis.
\newblock In \emph{European conference on computer vision}, pp.\  405--421.
  Springer, 2020.

\bibitem[Nguyen-Phuoc et~al.(2019)Nguyen-Phuoc, Li, Theis, Richardt, and
  Yang]{hologan}
Thu Nguyen-Phuoc, Chuan Li, Lucas Theis, Christian Richardt, and Yong-Liang
  Yang.
\newblock Hologan: Unsupervised learning of 3d representations from natural
  images.
\newblock In \emph{Proceedings of the IEEE/CVF International Conference on
  Computer Vision}, pp.\  7588--7597, 2019.

\bibitem[Niemeyer \& Geiger(2021)Niemeyer and Geiger]{giraffe}
Michael Niemeyer and Andreas Geiger.
\newblock Giraffe: Representing scenes as compositional generative neural
  feature fields.
\newblock In \emph{Proceedings of the IEEE/CVF Conference on Computer Vision
  and Pattern Recognition}, pp.\  11453--11464, 2021.

\bibitem[Noguchi et~al.(2021)Noguchi, Sun, Lin, and Harada]{noguchi2021neural}
Atsuhiro Noguchi, Xiao Sun, Stephen Lin, and Tatsuya Harada.
\newblock Neural articulated radiance field.
\newblock In \emph{Proceedings of the IEEE/CVF International Conference on
  Computer Vision}, pp.\  5762--5772, 2021.

\bibitem[Noguchi et~al.(2022)Noguchi, Sun, Lin, and Harada]{enarfgan}
Atsuhiro Noguchi, Xiao Sun, Stephen Lin, and Tatsuya Harada.
\newblock Unsupervised learning of efficient geometry-aware neural articulated
  representations.
\newblock \emph{arXiv preprint arXiv:2204.08839}, 2022.

\bibitem[Or-El et~al.(2022)Or-El, Luo, Shan, Shechtman, Park, and
  Kemelmacher-Shlizerman]{stylesdf}
Roy Or-El, Xuan Luo, Mengyi Shan, Eli Shechtman, Jeong~Joon Park, and Ira
  Kemelmacher-Shlizerman.
\newblock Stylesdf: High-resolution 3d-consistent image and geometry
  generation.
\newblock In \emph{Proceedings of the IEEE/CVF Conference on Computer Vision
  and Pattern Recognition}, pp.\  13503--13513, 2022.

\bibitem[Palafox et~al.(2021)Palafox, Bo{\v{z}}i{\v{c}}, Thies, Nie{\ss}ner,
  and Dai]{npms}
Pablo Palafox, Alja{\v{z}} Bo{\v{z}}i{\v{c}}, Justus Thies, Matthias
  Nie{\ss}ner, and Angela Dai.
\newblock Npms: Neural parametric models for 3d deformable shapes.
\newblock \emph{arXiv preprint arXiv:2104.00702}, 2021.

\bibitem[Palafox et~al.(2022)Palafox, Sarafianos, Tung, and
  Dai]{palafox2022spams}
Pablo Palafox, Nikolaos Sarafianos, Tony Tung, and Angela Dai.
\newblock Spams: Structured implicit parametric models.
\newblock In \emph{Proceedings of the IEEE/CVF Conference on Computer Vision
  and Pattern Recognition}, pp.\  12851--12860, 2022.

\bibitem[Pan et~al.(2020)Pan, Dai, Liu, Loy, and Luo]{2dganknows3d}
Xingang Pan, Bo~Dai, Ziwei Liu, Chen~Change Loy, and Ping Luo.
\newblock Do 2d gans know 3d shape? unsupervised 3d shape reconstruction from
  2d image gans.
\newblock \emph{arXiv preprint arXiv:2011.00844}, 2020.

\bibitem[Pavlakos et~al.(2019{\natexlab{a}})Pavlakos, Choutas, Ghorbani,
  Bolkart, Osman, Tzionas, and Black]{pavlakos2019expressive}
Georgios Pavlakos, Vasileios Choutas, Nima Ghorbani, Timo Bolkart, Ahmed~AA
  Osman, Dimitrios Tzionas, and Michael~J Black.
\newblock Expressive body capture: 3d hands, face, and body from a single
  image.
\newblock In \emph{Proceedings of the IEEE/CVF conference on computer vision
  and pattern recognition}, pp.\  10975--10985, 2019{\natexlab{a}}.

\bibitem[Pavlakos et~al.(2019{\natexlab{b}})Pavlakos, Choutas, Ghorbani,
  Bolkart, Osman, Tzionas, and Black]{smplx}
Georgios Pavlakos, Vasileios Choutas, Nima Ghorbani, Timo Bolkart, Ahmed~AA
  Osman, Dimitrios Tzionas, and Michael~J Black.
\newblock Expressive body capture: 3d hands, face, and body from a single
  image.
\newblock In \emph{Proceedings of the IEEE/CVF Conference on Computer Vision
  and Pattern Recognition}, pp.\  10975--10985, 2019{\natexlab{b}}.

\bibitem[Peng et~al.(2021{\natexlab{a}})Peng, Dong, Wang, Zhang, Shuai, Bao,
  and Zhou]{animatablenerf}
Sida Peng, Junting Dong, Qianqian Wang, Shangzhan Zhang, Qing Shuai, Hujun Bao,
  and Xiaowei Zhou.
\newblock Animatable neural radiance fields for human body modeling.
\newblock \emph{arXiv e-prints}, pp.\  arXiv--2105, 2021{\natexlab{a}}.

\bibitem[Peng et~al.(2021{\natexlab{b}})Peng, Zhang, Xu, Wang, Shuai, Bao, and
  Zhou]{neuralbody}
Sida Peng, Yuanqing Zhang, Yinghao Xu, Qianqian Wang, Qing Shuai, Hujun Bao,
  and Xiaowei Zhou.
\newblock Neural body: Implicit neural representations with structured latent
  codes for novel view synthesis of dynamic humans.
\newblock In \emph{Proceedings of the IEEE/CVF Conference on Computer Vision
  and Pattern Recognition}, pp.\  9054--9063, 2021{\natexlab{b}}.

\bibitem[Ranftl et~al.(2022)Ranftl, Lasinger, Hafner, Schindler, and
  Koltun]{Ranftl2022}
Ren\'{e} Ranftl, Katrin Lasinger, David Hafner, Konrad Schindler, and Vladlen
  Koltun.
\newblock Towards robust monocular depth estimation: Mixing datasets for
  zero-shot cross-dataset transfer.
\newblock \emph{IEEE Transactions on Pattern Analysis and Machine
  Intelligence}, 44\penalty0 (3), 2022.

\bibitem[Roich et~al.(2021)Roich, Mokady, Bermano, and
  Cohen-Or]{roich2021pivotal}
Daniel Roich, Ron Mokady, Amit~H Bermano, and Daniel Cohen-Or.
\newblock Pivotal tuning for latent-based editing of real images.
\newblock \emph{arXiv preprint arXiv:2106.05744}, 2021.

\bibitem[Sarkar et~al.(2021{\natexlab{a}})Sarkar, Golyanik, Liu, and
  Theobalt]{sarkar2021style}
Kripasindhu Sarkar, Vladislav Golyanik, Lingjie Liu, and Christian Theobalt.
\newblock Style and pose control for image synthesis of humans from a single
  monocular view.
\newblock \emph{arXiv preprint arXiv:2102.11263}, 2021{\natexlab{a}}.

\bibitem[Sarkar et~al.(2021{\natexlab{b}})Sarkar, Liu, Golyanik, and
  Theobalt]{humangan}
Kripasindhu Sarkar, Lingjie Liu, Vladislav Golyanik, and Christian Theobalt.
\newblock Humangan: A generative model of humans images.
\newblock \emph{arXiv preprint arXiv:2103.06902}, 2021{\natexlab{b}}.

\bibitem[Schwarz et~al.(2020)Schwarz, Liao, Niemeyer, and Geiger]{graf}
Katja Schwarz, Yiyi Liao, Michael Niemeyer, and Andreas Geiger.
\newblock Graf: Generative radiance fields for 3d-aware image synthesis.
\newblock \emph{Advances in Neural Information Processing Systems},
  33:\penalty0 20154--20166, 2020.

\bibitem[Shysheya et~al.(2019)Shysheya, Zakharov, Aliev, Bashirov, Burkov,
  Iskakov, Ivakhnenko, Malkov, Pasechnik, Ulyanov,
  et~al.]{shysheya2019textured}
Aliaksandra Shysheya, Egor Zakharov, Kara-Ali Aliev, Renat Bashirov, Egor
  Burkov, Karim Iskakov, Aleksei Ivakhnenko, Yury Malkov, Igor Pasechnik,
  Dmitry Ulyanov, et~al.
\newblock Textured neural avatars.
\newblock In \emph{Proceedings of the IEEE/CVF Conference on Computer Vision
  and Pattern Recognition}, pp.\  2387--2397, 2019.

\bibitem[Sitzmann et~al.(2020)Sitzmann, Martel, Bergman, Lindell, and
  Wetzstein]{sitzmann2020implicit}
Vincent Sitzmann, Julien Martel, Alexander Bergman, David Lindell, and Gordon
  Wetzstein.
\newblock Implicit neural representations with periodic activation functions.
\newblock \emph{Advances in Neural Information Processing Systems},
  33:\penalty0 7462--7473, 2020.

\bibitem[Skorokhodov et~al.(2022)Skorokhodov, Tulyakov, Wang, and
  Wonka]{epigraf}
Ivan Skorokhodov, Sergey Tulyakov, Yiqun Wang, and Peter Wonka.
\newblock Epigraf: Rethinking training of 3d gans.
\newblock \emph{arXiv preprint arXiv:2206.10535}, 2022.

\bibitem[Su et~al.(2021)Su, Yu, Zollh{\"o}fer, and Rhodin]{su2021nerf}
Shih-Yang Su, Frank Yu, Michael Zollh{\"o}fer, and Helge Rhodin.
\newblock A-nerf: Articulated neural radiance fields for learning human shape,
  appearance, and pose.
\newblock \emph{Advances in Neural Information Processing Systems},
  34:\penalty0 12278--12291, 2021.

\bibitem[Su et~al.(2022)Su, Bagautdinov, and Rhodin]{su2022danbo}
Shih-Yang Su, Timur Bagautdinov, and Helge Rhodin.
\newblock Danbo: Disentangled articulated neural body representations via graph
  neural networks.
\newblock \emph{arXiv preprint arXiv:2205.01666}, 2022.

\bibitem[Tewari et~al.(2021)Tewari, Thies, Mildenhall, Srinivasan, Tretschk,
  Wang, Lassner, Sitzmann, Martin-Brualla, Lombardi, et~al.]{advancesinnerf}
Ayush Tewari, Justus Thies, Ben Mildenhall, Pratul Srinivasan, Edgar Tretschk,
  Yifan Wang, Christoph Lassner, Vincent Sitzmann, Ricardo Martin-Brualla,
  Stephen Lombardi, et~al.
\newblock Advances in neural rendering.
\newblock \emph{arXiv preprint arXiv:2111.05849}, 2021.

\bibitem[Tsuchida et~al.(2019)Tsuchida, Fukayama, Hamasaki, and Goto]{aist}
Shuhei Tsuchida, Satoru Fukayama, Masahiro Hamasaki, and Masataka Goto.
\newblock Aist dance video database: Multi-genre, multi-dancer, and
  multi-camera database for dance information processing.
\newblock In \emph{Proceedings of the 20th International Society for Music
  Information Retrieval Conference, {ISMIR} 2019}, pp.\  501--510, Delft,
  Netherlands, November 2019.

\bibitem[Wang et~al.(2022)Wang, Schwarz, Geiger, and Tang]{wang2022arah}
Shaofei Wang, Katja Schwarz, Andreas Geiger, and Siyu Tang.
\newblock Arah: Animatable volume rendering of articulated human sdfs.
\newblock In \emph{European conference on computer vision}, volume~4, 2022.

\bibitem[Weng et~al.(2022)Weng, Curless, Srinivasan, Barron, and
  Kemelmacher-Shlizerman]{weng2022humannerf}
Chung-Yi Weng, Brian Curless, Pratul~P Srinivasan, Jonathan~T Barron, and Ira
  Kemelmacher-Shlizerman.
\newblock Humannerf: Free-viewpoint rendering of moving people from monocular
  video.
\newblock In \emph{Proceedings of the IEEE/CVF Conference on Computer Vision
  and Pattern Recognition}, pp.\  16210--16220, 2022.

\bibitem[Xiang et~al.(2022)Xiang, Yang, Deng, and Tong]{GRAMHD}
Jianfeng Xiang, Jiaolong Yang, Yu~Deng, and Xin Tong.
\newblock Gram-hd: 3d-consistent image generation at high resolution with
  generative radiance manifolds.
\newblock \emph{arXiv preprint arXiv:2206.07255}, 2022.

\bibitem[Xu et~al.(2021)Xu, Alldieck, and Sminchisescu]{xu2021h}
Hongyi Xu, Thiemo Alldieck, and Cristian Sminchisescu.
\newblock H-nerf: Neural radiance fields for rendering and temporal
  reconstruction of humans in motion.
\newblock \emph{Advances in Neural Information Processing Systems}, 34, 2021.

\bibitem[Yifan et~al.(2022)Yifan, Rahmann, and
  Sorkine-hornung]{yifan2022geometryconsistent}
Wang Yifan, Lukas Rahmann, and Olga Sorkine-hornung.
\newblock Geometry-consistent neural shape representation with implicit
  displacement fields.
\newblock In \emph{International Conference on Learning Representations}, 2022.
\newblock URL \url{https://openreview.net/forum?id=yhCp5RcZD7}.

\bibitem[Yoon et~al.(2021)Yoon, Liu, Golyanik, Sarkar, Park, and
  Theobalt]{yoon2021pose}
Jae~Shin Yoon, Lingjie Liu, Vladislav Golyanik, Kripasindhu Sarkar, Hyun~Soo
  Park, and Christian Theobalt.
\newblock Pose-guided human animation from a single image in the wild.
\newblock In \emph{Proceedings of the IEEE/CVF Conference on Computer Vision
  and Pattern Recognition}, pp.\  15039--15048, 2021.

\bibitem[Zablotskaia et~al.(2019)Zablotskaia, Siarohin, Zhao, and
  Sigal]{ubcfashion}
Polina Zablotskaia, Aliaksandr Siarohin, Bo~Zhao, and Leonid Sigal.
\newblock Dwnet: Dense warp-based network for pose-guided human video
  generation.
\newblock \emph{arXiv preprint arXiv:1910.09139}, 2019.

\bibitem[Zhang et~al.(2022)Zhang, Jiang, Yang, Xu, Shi, Song, Xu, Wang, and
  Feng]{avatargen}
Jianfeng Zhang, Zihang Jiang, Dingdong Yang, Hongyi Xu, Yichun Shi, Guoxian
  Song, Zhongcong Xu, Xinchao Wang, and Jiashi Feng.
\newblock Avatargen: a 3d generative model for animatable human avatars.
\newblock \emph{arXiv preprint arXiv:2208.00561}, 2022.

\bibitem[Zhang et~al.(2021)Zhang, Sangineto, Tang, Siarohin, Zhong, Sebe, and
  Wang]{3dawaresemanticgan}
Jichao Zhang, Enver Sangineto, Hao Tang, Aliaksandr Siarohin, Zhun Zhong, Nicu
  Sebe, and Wei Wang.
\newblock 3d-aware semantic-guided generative model for human synthesis.
\newblock \emph{arXiv preprint arXiv:2112.01422}, 2021.

\bibitem[Zhao et~al.(2021)Zhao, Yang, Zhang, Lin, Zhang, Yu, and
  Xu]{zhao2021humannerf}
Fuqiang Zhao, Wei Yang, Jiakai Zhang, Pei Lin, Yingliang Zhang, Jingyi Yu, and
  Lan Xu.
\newblock Humannerf: Generalizable neural human radiance field from sparse
  inputs.
\newblock \emph{arXiv preprint arXiv:2112.02789}, 2021.

\bibitem[Zheng et~al.(2021)Zheng, Yu, Liu, and Dai]{zheng2021pamir}
Zerong Zheng, Tao Yu, Yebin Liu, and Qionghai Dai.
\newblock Pamir: Parametric model-conditioned implicit representation for
  image-based human reconstruction.
\newblock \emph{IEEE transactions on pattern analysis and machine
  intelligence}, 44\penalty0 (6):\penalty0 3170--3184, 2021.

\end{thebibliography}
\bibliographystyle{iclr2023_conference}

\clearpage
\appendix\section{Appendix}

This is the supplementary material for \nickname{}. We provide more information of the training datasets in Sec.~\ref{sec:dataset}. More implementation details are introduced in Sec.~\ref{sec:imp}. More visual results and comparisons are provided in Sec.~\ref{sec:vis}. We also attached a demo video for better viewing experience.

\subsection{Datasets} \label{sec:dataset}

\noindent\textbf{DeepFashion}~\citep{deepfashion} collects fashion images from the internet. We only use images that contain the full body and not wearing dresses, which results in 8,036 images for training. We use SMPLify-X~\citep{smplx} to estimate SMPL parameters and camera parameters. All images are resized to $512\times 256$ for training. The alignment of the human body is the same as that proposed by \citet{jiang2022text2human}.

\begin{wrapfigure}{r}{0.3\textwidth}
\vspace{-20pt}
  \begin{center}
    \includegraphics[width=0.3\textwidth]{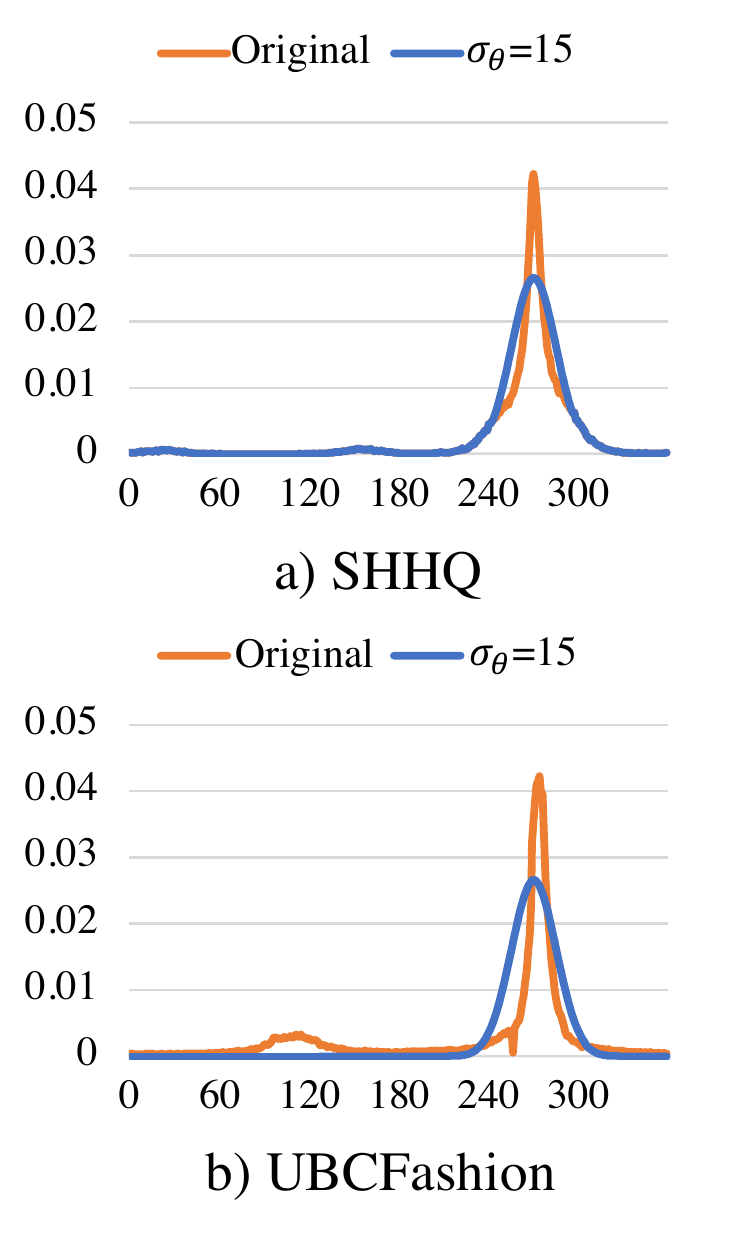}
  \end{center}
  \vspace{-10pt}
  \caption{Head Angle Distribution of SHHQ and UBCFashion.}
  \label{fig:supp_dist}
\end{wrapfigure}

\noindent\textbf{SHHQ}~\citep{styleganhuman} collects a larger-scale fashion dataset from the internet. It is processed similarly as DeepFashion, which results in 120,865 images in resolutions of $512\times 256$. In our experiments, we find that models trained using SMPL estimated by SMPLify-X performs better than that of SPIN~\citep{SPIN}. The head direction distribution of SHHQ, like DeepFashion, is also heavily imbalanced, as shown in Fig.~\ref{fig:supp_dist} a). The accompanying blue line is the distribution of the proposed pose-guided sampling.

\noindent\textbf{UBCFashion}~\citep{ubcfashion} is a fashion video dataset containing 500 sequences of models posing in front of the camera. Most models wear dresses in this dataset. We estimate SMPL sequences from videos by VIBE~\citep{kocabas2020vibe}. We use all frames of the 500 videos and crop them to $512\times256$ for training, which leads to 192,179 samples. Although most models spin in front of the camera, the head direction of UBCFashion is still heavily imbalanced, as shown in Fig.~\ref{fig:supp_dist} b).

\noindent\textbf{AIST}~\citep{aist} is a multi-view human dancing video dataset that provides rich poses and accurate SMPL estimations. We directly use the dataset processing scripts provided by ENARF-GAN~\citep{enarfgan} and get 72,000 samples. Each sample is resized to $256\times256$ for training.

\subsection{Implementation Details} \label{sec:imp}

\subsubsection{Network Architecture}

\begin{figure}
    \begin{center}
    \includegraphics[width=\linewidth]{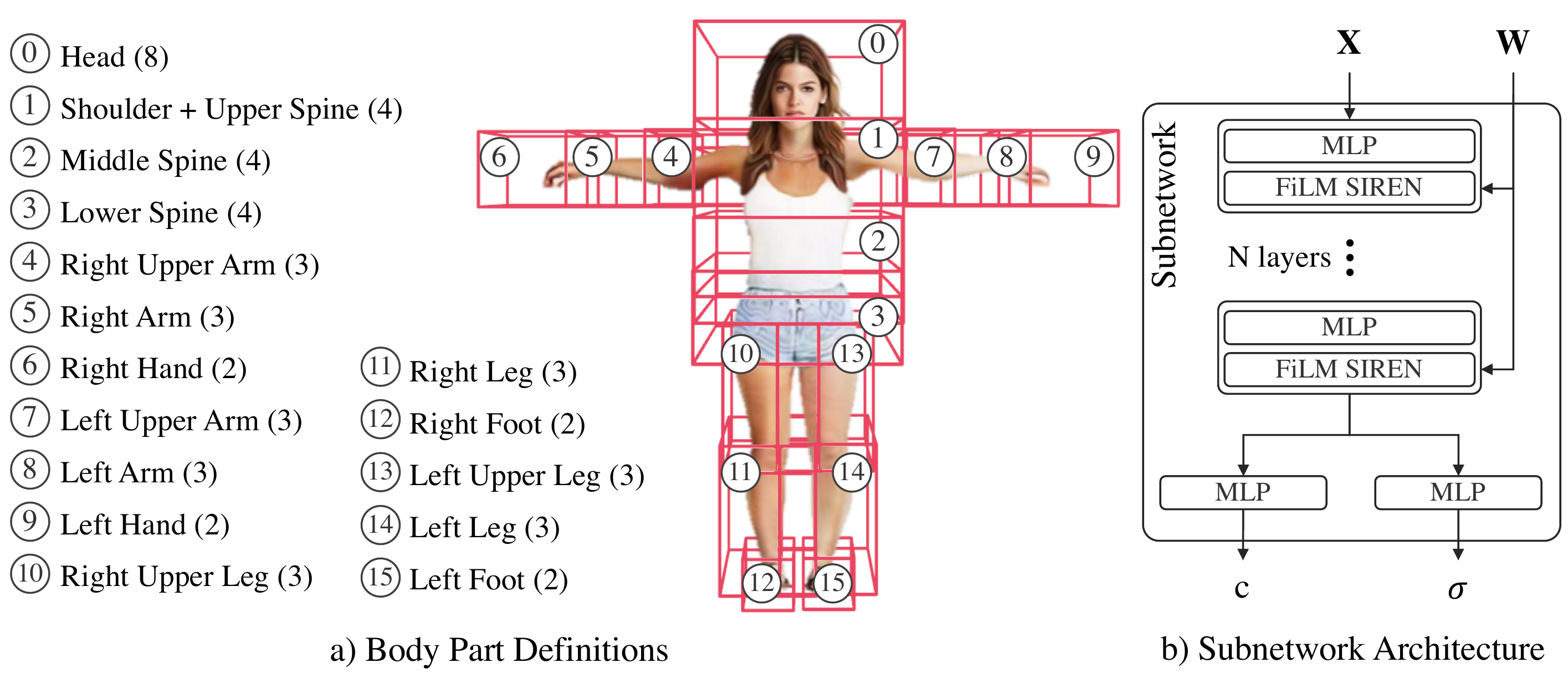}
    \end{center}
    \vspace{-12pt}
    \caption{a) shows our definition of 16 parts of human body. The number in the bracket is the number of corresponding subnetwork layers. b) shows the architecture of each subnetwork corresponding to each body part.}
    \label{fig:supp_network}
\end{figure}

As introduced in the main paper, we split the whole body into 16 parts, which is shown in Fig.~\ref{fig:supp_network} in specific. For each part, a subnetwork is assigned, which is developed based on StyleSDF~\citep{stylesdf}. The architecture of each subnetwork is shown in Fig.~\ref{fig:supp_network} b). For each subnetwork, multiple MLP and FiLM SIREN~\citep{pigan} activation layers are stacked alternatively. At the end of each subnetwork, two branches are used to separately estimate SDF value and RGB value. We assign different numbers of network layers for different body part empirically. Specific numbers are listed on Fig.~\ref{fig:supp_network} a). For the discriminator, we use the same architecture as that of StyleSDF~\citep{stylesdf}.

\subsubsection{Training Settings}

\noindent\textbf{Hyperparameters.} We use Adam optimizer~\citep{kingma2014adam} for the optimization of both generator and discriminator. The learning rate for generator is $2\times10^{-5}$. The learning rate for discriminator is $2\times10^{-4}$. The loss weights as set empirically as $\lambda_{\text{off}} = 1.5$ and $\lambda_{\text{eik}} = 0.5$. For one ray, 28 query points are sampled. For the pose-guided sample, we choose to use $\sigma_\theta = 15^{\circ}$.

\noindent\textbf{R1 Scheduler.} R1 regularization is used during training to penalize gradients of discriminator. Because it is highly challenging for the generator to learn plausible human appearance, the discriminator tends to overfit quickly if low R1 is set. But too high of R1 value would harm the final generation quality. Therefore, we set a R1 scheduler empirically, where R1 decrease from 300 to 18.5. R1 is cut in half every 50,000 iterations.

\noindent\textbf{Augmentation.} Inevitably, the SMPL estimations for 2D human images are not accurate for most samples. To compensate for the estimation error, we adopt small augmentations on real and fake samples before sent to the discriminator. The augmentation includes random panning, scaling and rotation in small ranges.

\noindent\textbf{Runtime Analysis.} The models are trained on 8 NVIDIA V100 GPUs for 5 days, with a batch size of 8. At test time, our model runs at $\sim5$ FPS on one NVIDIA V100 GPU.

\subsection{More Qualitative Results} \label{sec:vis}

\begin{figure}
    \begin{center}
    \includegraphics[width=\linewidth]{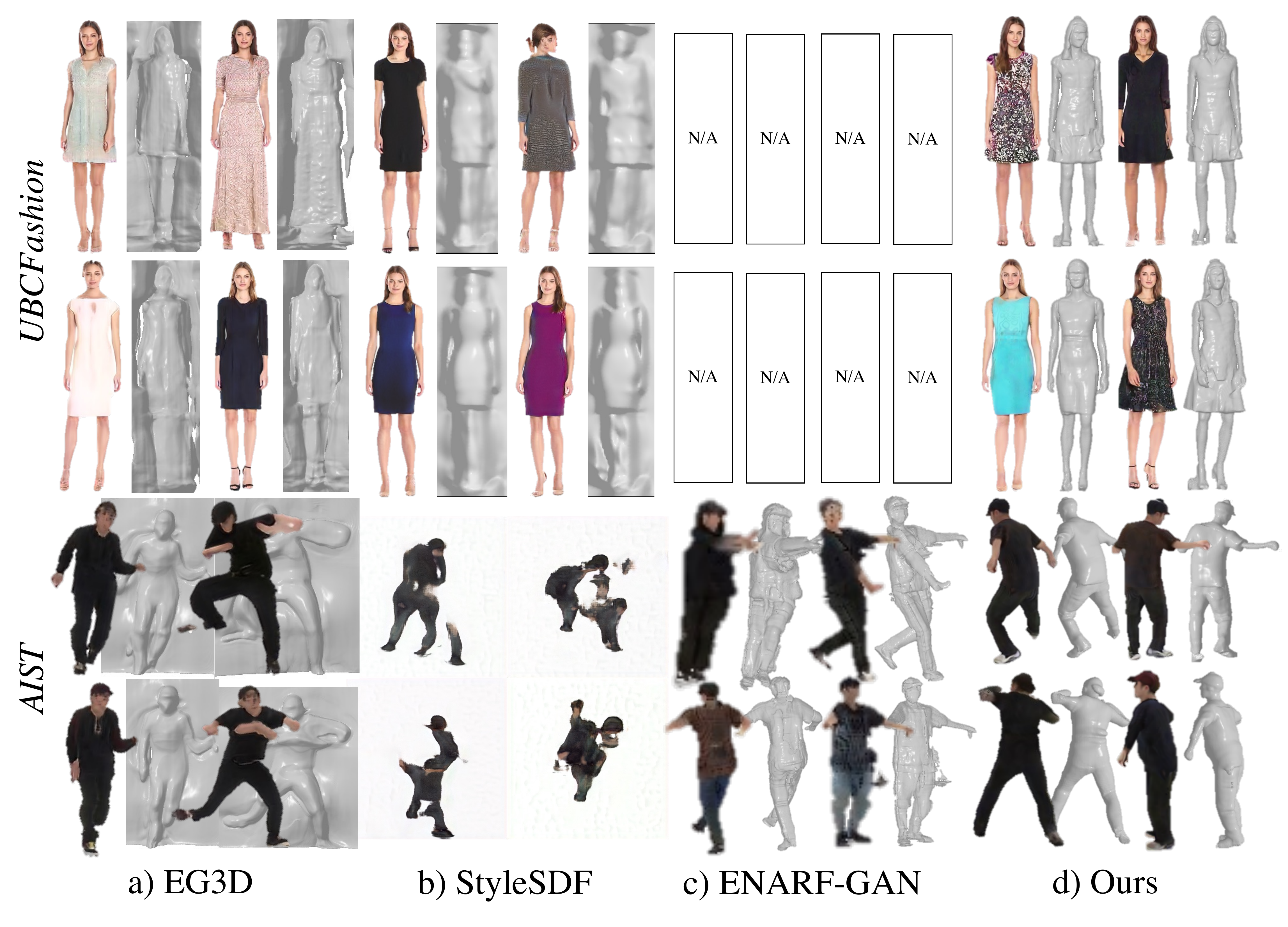}
    \end{center}
    \vspace{-12pt}
    \caption{Visual Comparison on UBCFashion \& AIST. Zoom in for the best view.}
    \label{fig:supp_comp}
\end{figure}

\noindent\textbf{Visual Comparison on UBCFashion \& AIST.} We further show renderings and corresponding meshes of three baseline methods and \nickname{} trained on UBCFashion and AIST in Fig.~\ref{fig:supp_comp}. UBCFashion has dense views and simple human poses. Therefore, EG3D and StyleSDF succeeded in generating reasonable renderings. But the corresponding meshes lack details due to training at low native resolution ($64\times64$). \nickname{} gives the best visual results among the baseline methods and also generates plausible meshes with reasonable details. Due to complex human poses, StyleSDF fails on AIST. EG3D manages to generate reasonable 3D human, but fails to capture correct human structure in some cases. ENARF-GAN, for its low-resolution training, loses most details and generates rough meshes. \nickname{} not only gets the best RGB renderings, but also generates meshes that preserve details like brims.

\begin{figure}
    \begin{center}
    \includegraphics[width=\linewidth]{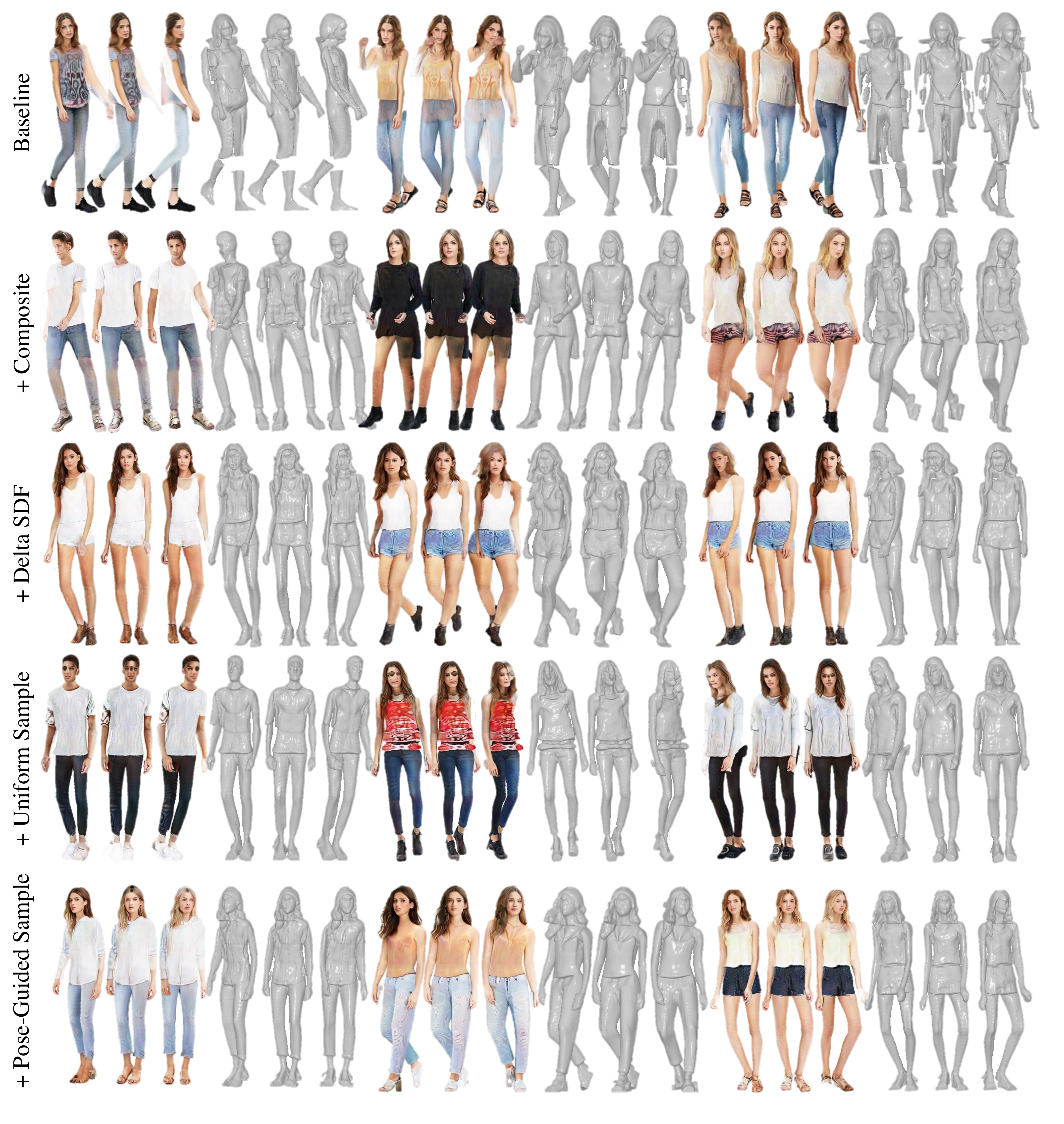}
    \end{center}
    \vspace{-12pt}
    \caption{Qualitative Evaluations on Ablation Studies. Zoom in for the best view.}
    \label{fig:supp_ablation}
\end{figure}

\noindent\textbf{Qualitative Evaluations on Ablation Studies.} As shown in Fig.~\ref{fig:supp_ablation}, we visualize renderings and geometry generated by baseline methods described in the ablation studies in the main paper. The ``Baseline'', due to being trained at lower resolution ($256\times128$), generates blurry renderings. The geometry fails to capture correct human structure (see broken knees). The compositional 3D human representation (``+ Composite'') facilitates high resolution training ($512\times256$). But lack of human prior leads to low-quality geometry (see unreasonable ``wrinkles'' on the upper bodies). By introducing a 3D human template and predicting delta SDF (``+ Delta SDF''), the visual quality increases and the geometry is mostly reasonable. However, the facial area is still flat due to the highly imbalanced viewing angle distribution. By using the pose-guided sampling (``+ Pose-Guided Sample''), we alleviate the imbalance issue and generate both high fidelity renderings and plausible geometry. To further validate our choice of Gaussian distribution in the pose-guided sampling, we visualize the results of models trained using uniform distribution (``+ Uniform Sample''). The middle of generated heads have severe artifacts.

\begin{figure}
    \begin{center}
    \includegraphics[width=\linewidth]{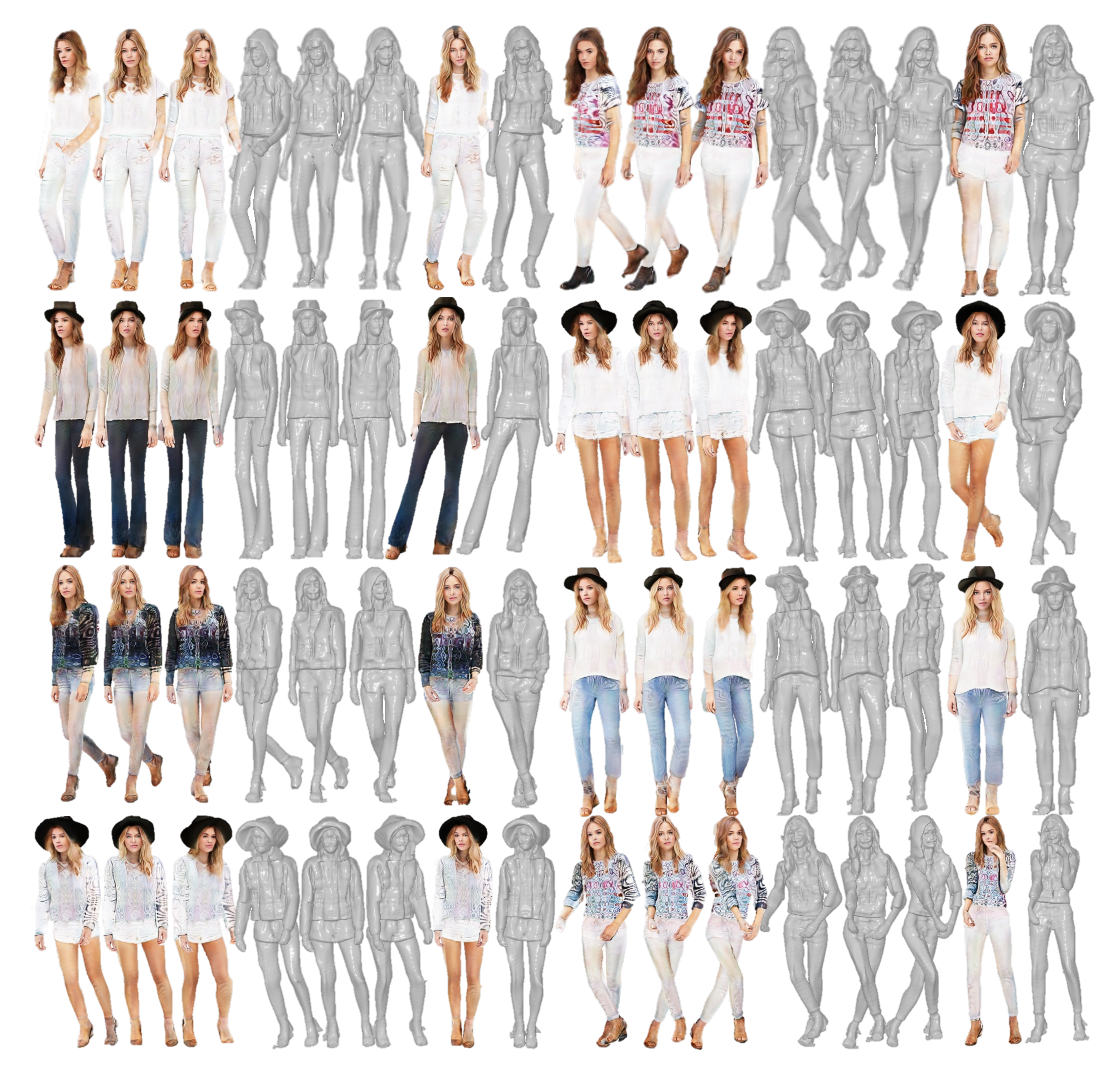}
    \end{center}
    \vspace{-12pt}
    \caption{More Qualitative Results of \nickname{} on DeepFashion. Zoom in for the best view.}
    \label{fig:supp_deepfashion}
\end{figure}

\begin{figure}
    \begin{center}
    \includegraphics[width=\linewidth]{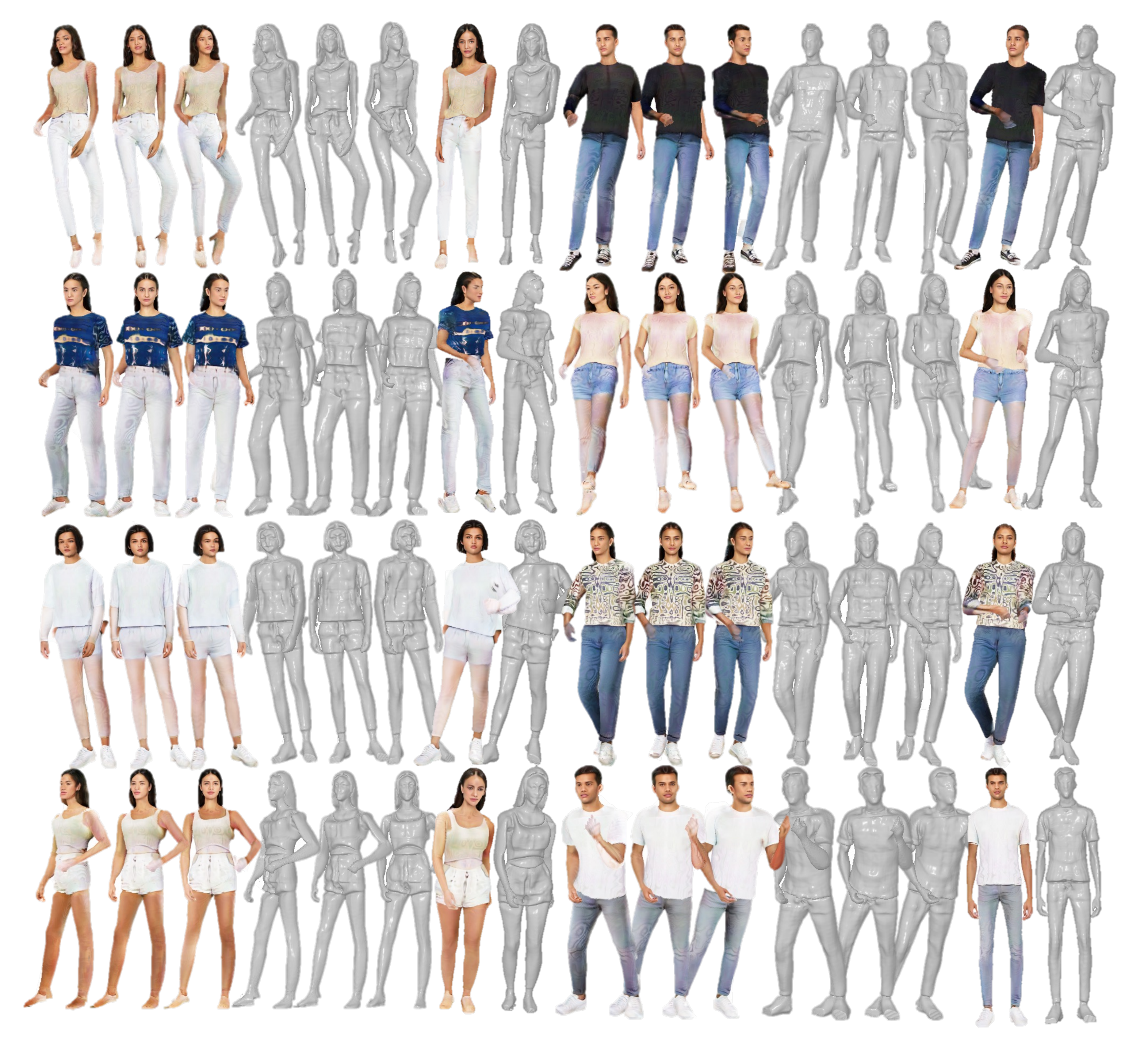}
    \end{center}
    \vspace{-12pt}
    \caption{More Qualitative Results of \nickname{} on SHHQ. Zoom in for the best view.}
    \label{fig:supp_20wfashion}
\end{figure}

\begin{figure}
    \begin{center}
    \includegraphics[width=\linewidth]{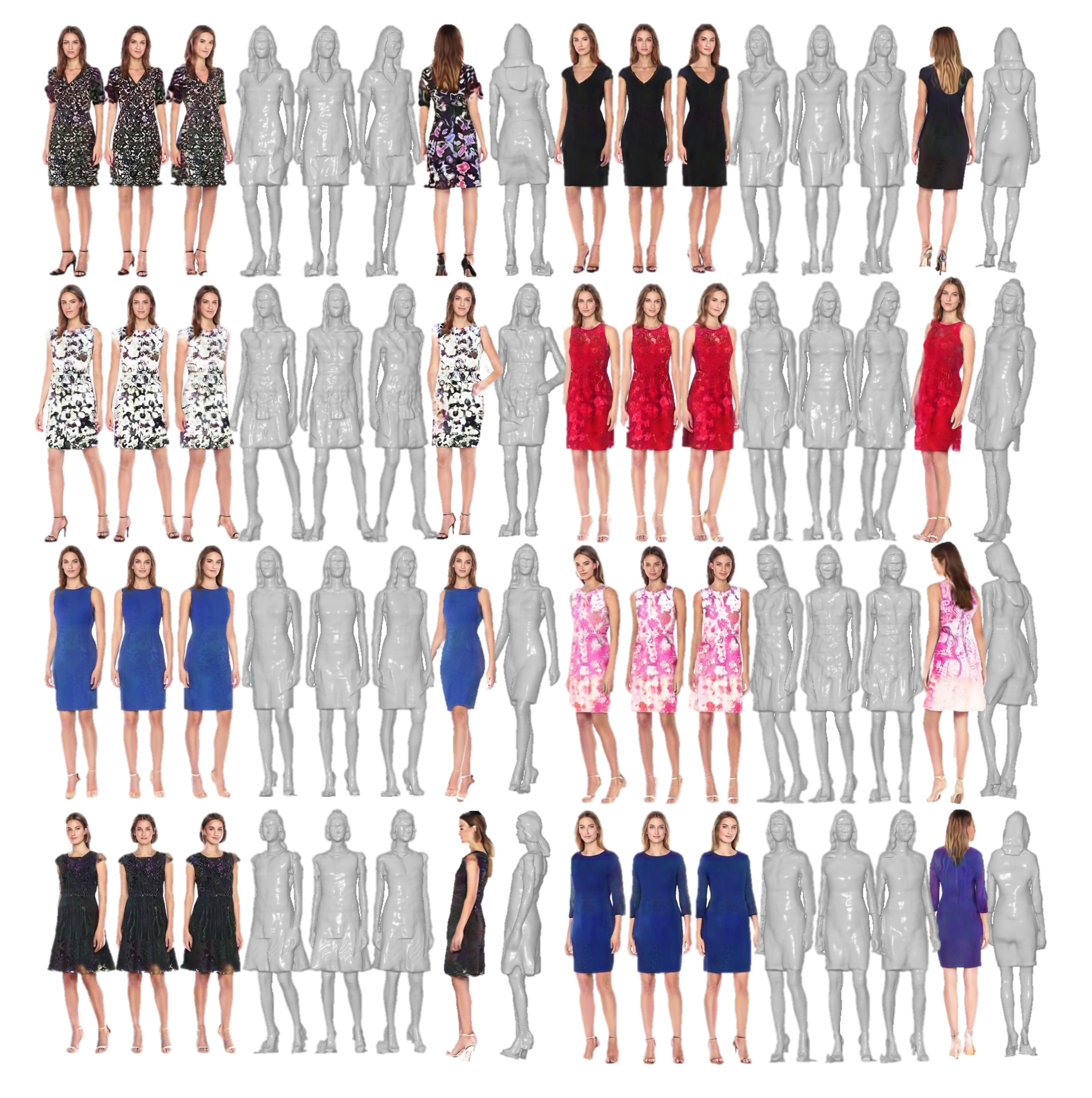}
    \end{center}
    \vspace{-12pt}
    \caption{More Qualitative Results of \nickname{} on UBCFashion. Zoom in for the best view.}
    \label{fig:supp_ubcfashion}
\end{figure}

\begin{figure}
    \begin{center}
    \includegraphics[width=\linewidth]{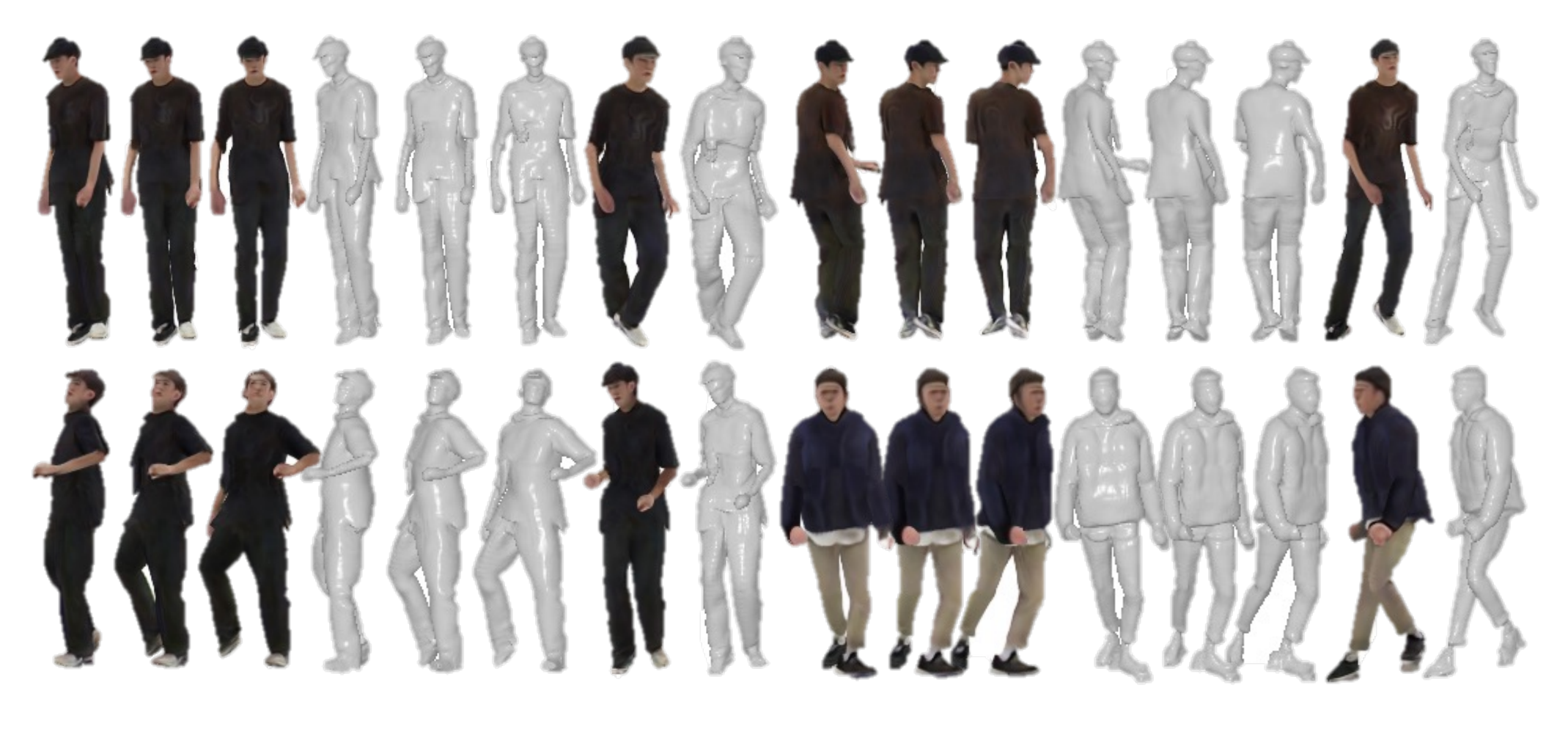}
    \end{center}
    \vspace{-12pt}
    \caption{More Qualitative Results of \nickname{} on AIST. Zoom in for the best view.}
    \label{fig:supp_aist}
\end{figure}

\noindent\textbf{More Qualitative Results of \nickname{}.} More qualitative results of \nickname{} on four datasets are shown in Fig.~\ref{fig:supp_deepfashion}, \ref{fig:supp_20wfashion}, \ref{fig:supp_ubcfashion}, \ref{fig:supp_aist}. For each sample, we show its novel view renderings and novel pose rendering.

\end{document}